\documentclass[pmlr,twocolumn,10pt]{jmlr} %
\makeatletter
\renewcommand{\@titlefoot}{}
\makeatother

\makeatletter
\providecommand{\mlhtrack}[1]{\def\mlhtrackname{#1}}
\@ifundefined{ifmlhneedspmlr}{\newif\ifmlhneedspmlr}{}
\@ifundefined{ifmlhfindings}{\newif\ifmlhfindings}{}
\@ifundefined{ifmlhdemo}{\newif\ifmlhdemo}{}
\makeatother

\mlhtrack{proceedings}

\newif\iffinal
\finalfalse  %

\iffinal
    \ifmlhneedspmlr
      \jmlrvolume{XXX}
      \jmlryear{2026}
    \fi
    \ifmlhfindings \jmlrproceedings{}{ML4H 2026 - Findings Track}\fi
    \ifmlhdemo     \jmlrproceedings{}{ML4H 2026 - Demo Track}\fi
    \jmlrworkshop{Machine Learning for Health (ML4H) 2026}
\else
    \jmlrproceedings{}{Accepted at Machine Learning for Health (ML4H) 2026}
    \jmlrworkshop{}
\fi

\usepackage{booktabs}
\usepackage{array}
\usepackage{siunitx}

\usepackage[switch]{lineno}

\theorembodyfont{\upshape}
\theoremheaderfont{\scshape}
\theorempostheader{:}
\theoremsep{\newline}

\newcommand{\frozenicon}{\raisebox{-0.16em}{\includegraphics[height=1.16em]{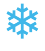}}}
\newcommand{\finetunedicon}{\raisebox{-0.14em}{\includegraphics[height=1.05em]{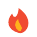}}}
\usepackage{colortbl}
\usepackage{placeins}
\usepackage{float}

\definecolor{hlblue}{rgb}{0.878,0.941,0.988}
\definecolor{sdteal}{rgb}{0.16,0.52,0.45}

\newcommand{\sdc}[2]{%
  \mbox{#1\,{\scriptsize\color{sdteal}$\pm$#2}}%
}

\usepackage{xcolor}
\usepackage{soul}
\usepackage[most]{tcolorbox}
\usepackage{cuted}
\usepackage{capt-of}
\usepackage{needspace}
\definecolor{suppboxheader}{HTML}{454545}
\definecolor{suppboxframe}{HTML}{686868}
\newtcolorbox{supplementbox}[2][]{
  enhanced,
  breakable,
  colback=white,
  colframe=suppboxframe,
  colbacktitle=suppboxheader,
  coltitle=white,
  fonttitle=\bfseries,
  title={#2},
  boxrule=0.7pt,
  arc=1.5mm,
  left=3mm,
  right=3mm,
  top=2.5mm,
  bottom=2.5mm,
  before skip=3pt,
  after skip=2pt,
  #1
}

\title[Language Models for Personalized Sleep Guidance]%
{Toward Personalized Sleep Guidance from Wearable Data Using Language Models}

\author{\centering
\Name{Yusheng Tan\nametag{$^{1,*}$}},
\Name{Running Zhao\nametag{$^{1,*}$}},
\Name{Sofia Angel\nametag{$^{1}$}},
\Name{Ninghui Hao\nametag{$^{1}$}},
\Name{Ash Arian\nametag{$^{1}$}},
\Name{Nikita N. Dulin\nametag{$^{1}$}},
\Name{Jay Lin\nametag{$^{1}$}},
\Name{Ou Zhu\nametag{$^{1}$}},
\Name{Faiza Shaik\nametag{$^{1}$}},
\Name{Xinxing Yang\nametag{$^{1}$}},
\Name{Bonnie W. Leung\nametag{$^{2}$}},
\Name{Katie Roster\nametag{$^{3}$}},
\Name{Arlene {Ruiz de Luzuriaga}\nametag{$^{1}$}},
\Name{Kenneth Lee\nametag{$^{1}$}},
\Name{Alejandra Lastra\nametag{$^{1}$}},
\Name{Habibul Ahsan\nametag{$^{1}$}},
\Name{Guihong C. Wan\nametag{$^{1,\dagger}$}}\\
\addr $^{1}$The University of Chicago\\
$^{2}$Baylor College of Medicine\\
$^{3}$Georgetown University School of Medicine\\
$^{*}$Equal contribution\\
$^{\dagger}$Corresponding author: gwan@uchicago.edu\\
}

\begin{document}

\maketitle
\raggedbottom   

\begin{abstract}
Sleep monitoring using wearable data has shown promise for personal health, yet large language model (LLM)-based summarization and question answering remain insufficient for personalized sleep guidance. 
Training specialized models, however, 
often requires costly expert annotation. 
Moreover, privacy and accessibility concerns motivate lightweight, local deployment for end users. 
We present a two-stage framework to address these challenges. 
Specifically, in Stage~1, a multi-agent LLM pipeline reasons structured sleep guidance from unannotated wearable records, enabling scalable dataset construction.
Stage~2 distills
guidance reasoning trajectories
into small language models (SLMs) through supervised fine-tuning and integrates a training-free Best-of-$N$ selection strategy to enhance inference.
Experimental results demonstrate our method outperforms commercial general and medical LLMs and open-source models.
Human evaluation 
further supports the quality of the generated guidance and the feasibility of personalized sleep guidance with SLMs.

\end{abstract}


\begin{figure}[!t]
    \centering
    \includegraphics[width=\columnwidth,
    trim=0.5cm 0 5.5cm 0cm,
    clip]{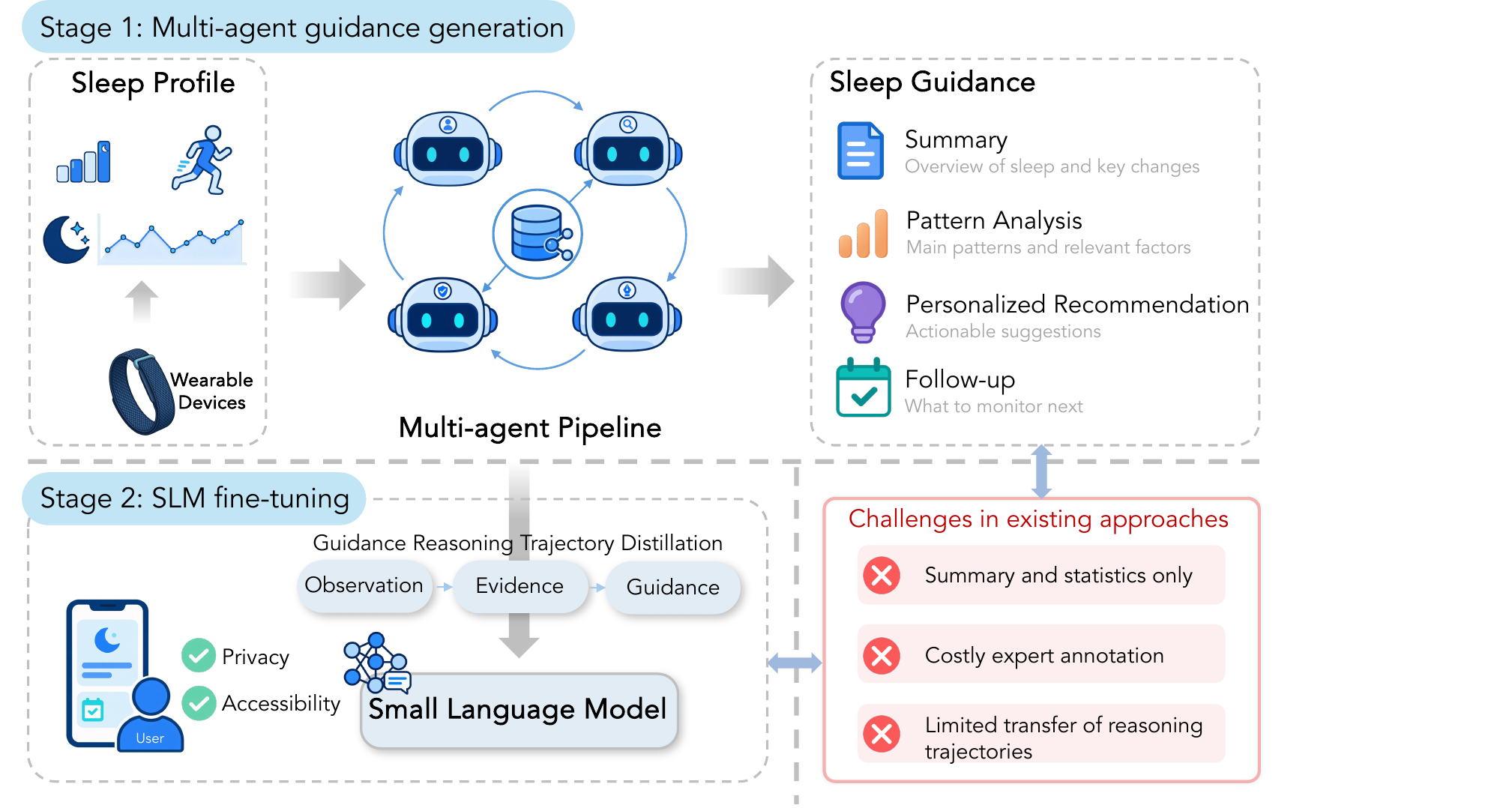}
    \vspace{-14pt}
    \caption{\textbf{Overview of our framework for personalized sleep guidance.}}
    \label{fig:teaser}
    \vspace{-12pt}
\end{figure}

\paragraph*{Data and Code Availability} The data used in this study are available to approved researchers through the All of Us Researcher Workbench upon registration, completion of ethics training, and attestation to the Data Use Agreement \citep{allofus2026cdr9}. The code is available at \url{https://github.com/cwanlab/Personalized-Sleep-Guidance}.

\paragraph*{Institutional Review Board (IRB)} This work uses the All of Us dataset \citep{allofus2026cdr9} that is accessible only to approved researchers. Human evaluation was provided by the research team. IRB review was not required.

\section{Introduction}
\label{sec:intro}
Sleep is closely related to overall health and well-being. Persistent sleep problems
are associated with a broad range of adverse health outcomes \citep{buysse2014sleep,chung2021multidimensional}. With advances in wearable technology, consumer wearable devices have become increasingly integrated into everyday health monitoring. Recent nationally representative survey data show that wearable use for health tracking among U.S. adults increased from 30.2\% in 2020 to 41.1\% in 2024 \citep{pedroso2026wearable}. Wearable devices commonly combine motion sensing, such as accelerometry, with optical measurements, such as photoplethysmography, to estimate sleep-related measures during routine use \citep{imtiaz2021systematic}. The growing adoption of these devices makes personal sleep data increasingly accessible, creating an opportunity to provide personalized sleep guidance from routinely collected wearable data.

Recent studies have begun to use large language models (LLMs) to provide personalized
health support from wearable data. 
PhysioLLM \citep{fang2024physiollm}, PHIA \citep{merrill2026transforming}, and HealthGuru \citep{wang2025exploring} support data analysis,
question answering, and conversational interaction over personal wearable data, but primarily 
focus on
basic sleep data analysis and summary.
WEQA extends the user-initiated query answering with
query-adaptive agentic reasoning \citep{zhang2026weqa}. 
PH-LLM moves closer to personalized guidance by directly
generating insights and recommendations for sleep and fitness, but 
it relies on domain-expert-authored case studies
\citep{khasentino2025personal}.

These limitations, summarized in Figure~\ref{fig:teaser}, expose two challenges for personalized sleep guidance.
First, useful guidance should go beyond query-driven data analyses or summaries to 
identify
salient patterns across multiple nights and connect them with relevant knowledge to generate actionable, evidence-grounded recommendations. 
Second, expert annotation is resource-intensive and difficult to scale for personalized sleep guidance. 
Although
synthetic and LLM-generated guidance can reduce this burden, ensuring that automatically
constructed guidance is consistent with the original wearable records and supported by evidence remains challenging.

Practical deployment introduces a further challenge. Personalized sleep guidance involves sensitive personal wearable data, making local or on-device inference attractive to reduce reliance on external services \citep{aminifar2024privacy}. Recent work has begun to distill personalized sleep capabilities into small language models (SLMs) for lightweight and edge deployment \citep{zheng2025profile}. However, these approaches focus primarily on transferring basic sleep analysis and summary,
whereas explicitly modeling intermediate reasoning processes into SLMs, including characterization of multi-night sleep patterns, selection of relevant evidence, and formulation of recommendations, remains underexplored.

Our framework addresses these challenges in two stages, as illustrated in Figure~\ref{fig:teaser}. In Stage~1, we propose a multi-agent LLM pipeline to reason structured sleep guidance from
unannotated longitudinal wearable records.
Beyond basic sleep analysis and summary, specialized agents characterize pattern observations, select supporting evidence, formulate personalized guidance, and verify each case through deterministic checks and independent reverse review.
This process reduces reliance on case-by-case expert annotation while enabling scalable dataset construction.
In Stage~2, we distill the guidance reasoning trajectories into SLMs by fine-tuning on observations, evidence, and guidance generated from Stage~1.
To reduce error propagation from an observation, we introduce a training-free Best-of-$N$ strategy that selects a high-quality sampled observation
before 
evidence retrieval. 
This two-stage design decouples large-model construction from small-model deployment, enabling evidence-grounded guidance without rerunning the multi-agent
LLM pipeline during deployment, thereby enhancing privacy and accessibility. 

Experimental results illustrate the superior performance of our method across Qwen3.5 models with 0.8B, 2B, and 4B parameters. Notably, our 0.8B model outperforms the 4B baseline and commercial general and medical LLMs, including GPT-5.6-Sol, GPT Health, and OpenEvidence, across all metrics.
We further conducted stage-specific human evaluations of 
the sleep guidance structure (Stage~1) and
generated sleep guidance (Stage~2). 
Expert assessment 
supported
 the Stage~1
framework, with a mean rating of 4.67/5 across rubric items. 
Evaluation of our 0.8B model's outputs by MD students yielded average ratings of
4.99/5 for safety,
4.74/5 for readability, 
and 4.06/5 for overall quality. 
These results support both
the appropriateness of the proposed framework
 and the capability of SLMs for personalized sleep guidance.

Our main contributions are:
\begin{itemize}

\item We introduce a multi-agent LLM pipeline to derive structured sleep guidance from longitudinal wearable records, reducing reliance on costly expert annotation and enabling scalable dataset construction.

\item We distill 
personalized
guidance 
capability into SLMs through supervised fine-tuning on the reasoning trajectory and 
a training-free Best-of-$N$ strategy at inference.

\item We conduct comprehensive experiments across model scales, combining quantitative comparisons and human evaluations to demonstrate the superiority of our framework over baselines and the feasibility of SLMs for personalized sleep guidance.

\end{itemize}

\section{Related Work}
\label{sec:related}

\subsection{Large Language Models for Wearable Sleep Data}

Language models have increasingly been used to interpret personal 
wearable data. 
PhysioLLM \citep{fang2024physiollm}, 
PHIA \citep{merrill2026transforming}, and HealthGuru \citep{wang2025exploring}
support a range of health-related tasks,
including statistical analysis, tool-assisted reasoning, and question answering.
WEQA further introduces query-adaptive agentic reasoning for wearable health question answering \citep{zhang2026weqa}. Moving beyond query-driven analysis, PH-LLM generates personalized insights and recommendations for sleep and fitness
\citep{khasentino2025personal}. 
These studies demonstrate a progression toward personalized guidance, while explicit interpretation of multi-night patterns and selection of supporting evidence for evidence-grounded 
decision-making 
remain less explored.

\subsection{Scalable Training Data Construction}

Personalized sleep guidance also raises the question of how high-quality training data can be constructed at scale. PH-LLM relies on domain-expert-authored case studies \citep{khasentino2025personal}, whereas SleepCoT and subsequent work use synthetic or LLM-guided data construction to reduce manual effort \citep{zheng2025profile}. More broadly, Self-Instruct \citep{wang2023self} bootstraps instruction-response data from model generation,
while multiple agentic workflows \citep{mitra2024agentinstruct,chen2024sharegpt4video,noteit} are proposed to generate large-scale synthetic data. However, personalized wearable guidance additionally requires generated 
training examples
to remain consistent with individual records and traceable to the evidence supporting the guidance.

\subsection{Fine-Tuning with Intermediate Reasoning and Evidence}

Final-answer-only supervised fine-tuning learns from input-answer pairs without explicit supervision of intermediate reasoning,
whereas rationale-based distillation can embed the reasoning process into smaller models \citep{hsieh2023distilling}. 
Retrieval-augmented generation grounds model outputs in external knowledge \citep{lewis2020retrieval}, and generative retrieval methods can select supporting evidence by generating identifiers before response generation
\citep{sun2023generative}. 
These directions motivate combining intermediate
decision-making 
with external evidence. 
For personalized sleep guidance, however, longitudinal pattern recognition and relevant evidence selection have rarely been treated together as explicit, verifiable intermediate decisions before generating the final guidance.

\begin{figure*}[t]
\centering
\includegraphics[width=0.9\textwidth]{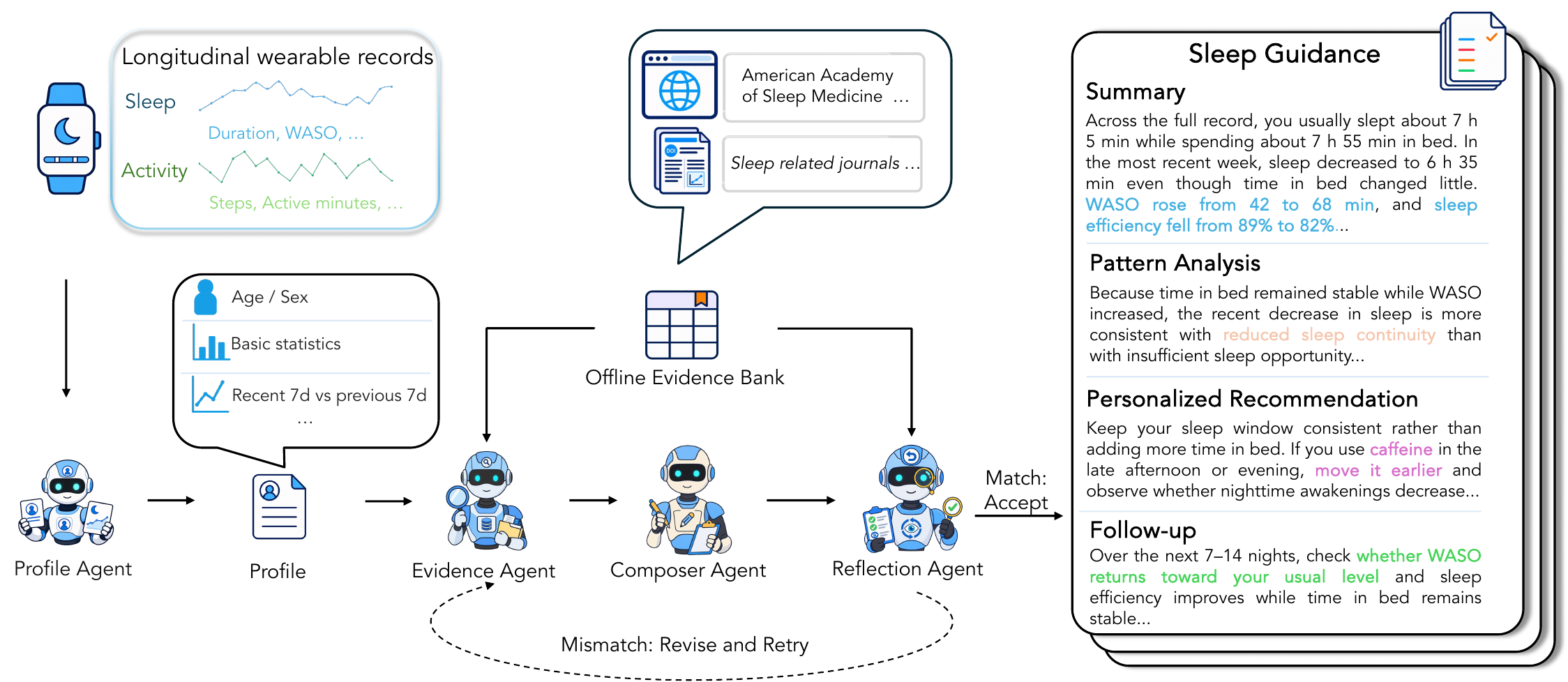}
\caption{Stage~1 multi-agent pipeline for deriving structured sleep guidance from wearable records.
Only examples passing deterministic checks and independent reverse review
are retained.}
\label{fig:stage1}
\end{figure*}

\section{Methods}
\label{sec:method}

\subsection{Dataset and Problem Formulation}
\label{sec:data}

We obtained Fitbit-derived sleep and physical activity records from $1{,}500$ adults aged 18 years or older through the Controlled Tier of the All of Us Researcher Workbench using Curated Data Repository C2025Q4R6 \citep{patten2026all, allofus2026cdr9}. We randomly sampled 750 female and 750 male participants (see Appendix~\ref{app:details} for the selection criteria). For each participant, we constructed one case containing age, sex, and 14--60 consecutive days of complete daily sleep and activity measurements, yielding $36{,}657$ daily records in total.
The 14-day minimum provides two non-overlapping seven-night windows for comparing
recent and preceding sleep patterns, while allowing sleep patterns to be characterized over multiple nights \citep{schutte2008clinical,krahn2021recommended}. Multi-night monitoring is particularly important for characterizing sleep variability beyond mean sleep measures \citep{bei2016beyond,lau2022minimum}. 
We cap each case at 60 days (approximately 8.5 weeks), 
within the 6--10-week range reported to reliably characterize intraindividual sleep variability, while bounding sequence length \citep{leota2026many}.  
We 
randomly
partitioned 
the dataset at the participant level into $1{,}200$ training cases and $300$ held-out test cases.

For case $i$, let $X_i$ denote the raw input data, consisting of age, sex, and
$T_i$ consecutive days of wearable measurements, where
$T_i \in [14,60]$. 
We construct a profile packet $P_i$, organizing 
daily measurements 
and 
deriving
basic 
statistics for subsequent analyses (see Appendix~\ref{app:details} for details).

Each case is associated with an observation $O_i$, which captures 
salient
multi-night sleep patterns,
and an ordered sequence of evidence identifiers
$E_i=(e_{i1},\ldots,e_{iK_i})$,
where $K_i$ is the number of evidence identifiers for case $i$.
The selected identifiers are used to retrieve the evidence cards $C_i$,
which provide supporting evidence for generating a four-section guidance $R_i$ (see Section~\ref{sec:stage1}).

Each evidence identifier refers to one of 22 evidence cards in the offline Evidence Bank $\mathcal{B}$. These cards were derived from 46 sources, including peer-reviewed sleep literature and authoritative health organizations and websites (see Appendix~\ref{app:details}). 
Given $E_i$, the corresponding evidence cards are
retrieved deterministically: 
$C_i=\mathcal B[E_i]$.

Stage~1 constructs the evidence-grounded dataset of 1,500 cases:
\begin{equation}
\begin{aligned}
\mathcal D^\star
&=\bigl\{(P_i,O_i^\star,E_i^\star,C_i^\star,R_i^\star)\bigr\}_{i=1}^{1{,}500}.
\end{aligned}
\label{eq:dataset}
\end{equation}
The ordered 
$P {\rightarrow} O {\rightarrow} E {\rightarrow} C {\rightarrow} R$ process defines the guidance reasoning trajectory. 
In Stage~2, we distill
this trajectory
into SLMs.

\subsection{Multi-Agent for Structured Sleep Guidance Generation}
\label{sec:stage1}

Stage~1 proposes a multi-agent LLM pipeline that transforms unannotated longitudinal wearable records into 
four-section sleep guidance $R_i$, reducing reliance on case-by-case expert annotation and enabling scalable dataset construction.
Each guidance 
$R_i$ 
follows 
the same four-section 
structure, including
\emph{Summary}, \emph{Pattern Analysis}, \emph{Personalized Recommendation},
and \emph{Follow-up}.

\emph{Summary} provides a concise, standalone description of the
wearable
records of a participant, including the 
general
sleep patterns, the most relevant
longitudinal change, and dated fluctuations when they clarify the primary sleep pattern. The summary
describes multiple dimensions of sleep health without introducing interpretation
or advice \citep{buysse2014sleep,chung2021multidimensional}.

\emph{Pattern Analysis} identifies the primary repeated pattern and the 
relevant
case-specific factors, then interprets the pattern using 
selected evidence. 
It focuses on explaining 
the primary pattern
without
introducing recommendations, unsupported causes, 
or diagnostic claims.

\emph{Personalized Recommendation} translates the pattern analysis
into feasible actions. Consistent with the \emph{Advise} and
\emph{Assist} components of the 5As Behavioral Counseling Framework
\citep{whitlock2002evaluating}, recommendations are tied to observed 
facts; factors
not contained in $P_i$ may appear only as explicit conditional possibilities.

\emph{Follow-up} specifies how the primary recommendation should be
reassessed over a comparable multi-night window using the same sleep dimension.
This design emphasizes
repeated measurements rather than interpretation of a single subsequent night
\citep{barber2023collect}.

Together, the four sections adapt the clinical reasoning cycle of information
collection, interpretation, action, and outcome evaluation
\citep{levett2010five} to evidence-grounded sleep guidance.

The construction pipeline uses LLM agents with specialized roles,
as shown in Figure~\ref{fig:stage1}.
The Profile Agent analyzes the raw wearable records
to construct the profile packet $P_i$,
generating and executing code to compute quantitative summaries and comparisons.
In cycle $k\in\{1,\ldots,5\}$, the Evidence Agent produces an observation
$O_i^{(k)}$ capturing 
salient multi-night sleep patterns and selects the
corresponding evidence identifiers $E_i^{(k)}$. The selected evidence cards
are retrieved verbatim as:
$
C_i^{(k)}=\mathcal B[E_i^{(k)}].
$
The Composer Agent uses $P_i$ and $C_i^{(k)}$ to generate the four-section
guidance $R_i^{(k)}$ and records numeric anchors $A_i^{(k)}$.
Deterministic checks verify the consistency of $A_i^{(k)}$ with the wearable records and selected evidence. 

Only outputs that pass the deterministic checks proceed to independent reverse review.
The Reflection Agent independently recovers the evidence required by
$R_i^{(k)}$ using $P_i$ and the 
full 
Evidence Bank $\mathcal B$, without
access to the evidence selected by the Evidence Agent. 
Let $\Gamma_i^{(k)}\in\{0,1\}$ denote the verification
outcome, with
$
\Gamma_i^{(k)}=1
$
only if the recovered evidence identifiers exactly match those
selected by the Evidence Agent. 

If $\Gamma_i^{(k)}=1$, the example is retained~as
$(P_i,O_i^\star,E_i^\star,C_i^\star,R_i^\star)$; otherwise, the reverse-review
result is used to revise the example in the next cycle. The process continues
until verification succeeds or \textcolor{black}{five cycles are completed}. 
This verification checks consistency with the wearable records and traceability to supporting evidence.
Appendices~\ref{app:details} and~\ref{app:composer-prompt} 
provide additional 
details.

\subsection{Guidance Reasoning Trajectory Distillation}
\label{sec:stage2}

Running the Stage~1 pipeline for a new 
case
would require
multiple LLM calls and repeated verification. To support lightweight local deployment,
we distill the guidance reasoning trajectory into SLMs through
supervised fine-tuning (SFT).
We use Qwen3.5 models with 0.8, 2, and 4 billion parameters and apply the same
training procedure across all models.

\begin{figure}[t]
\centering
\includegraphics[width=0.90\columnwidth]{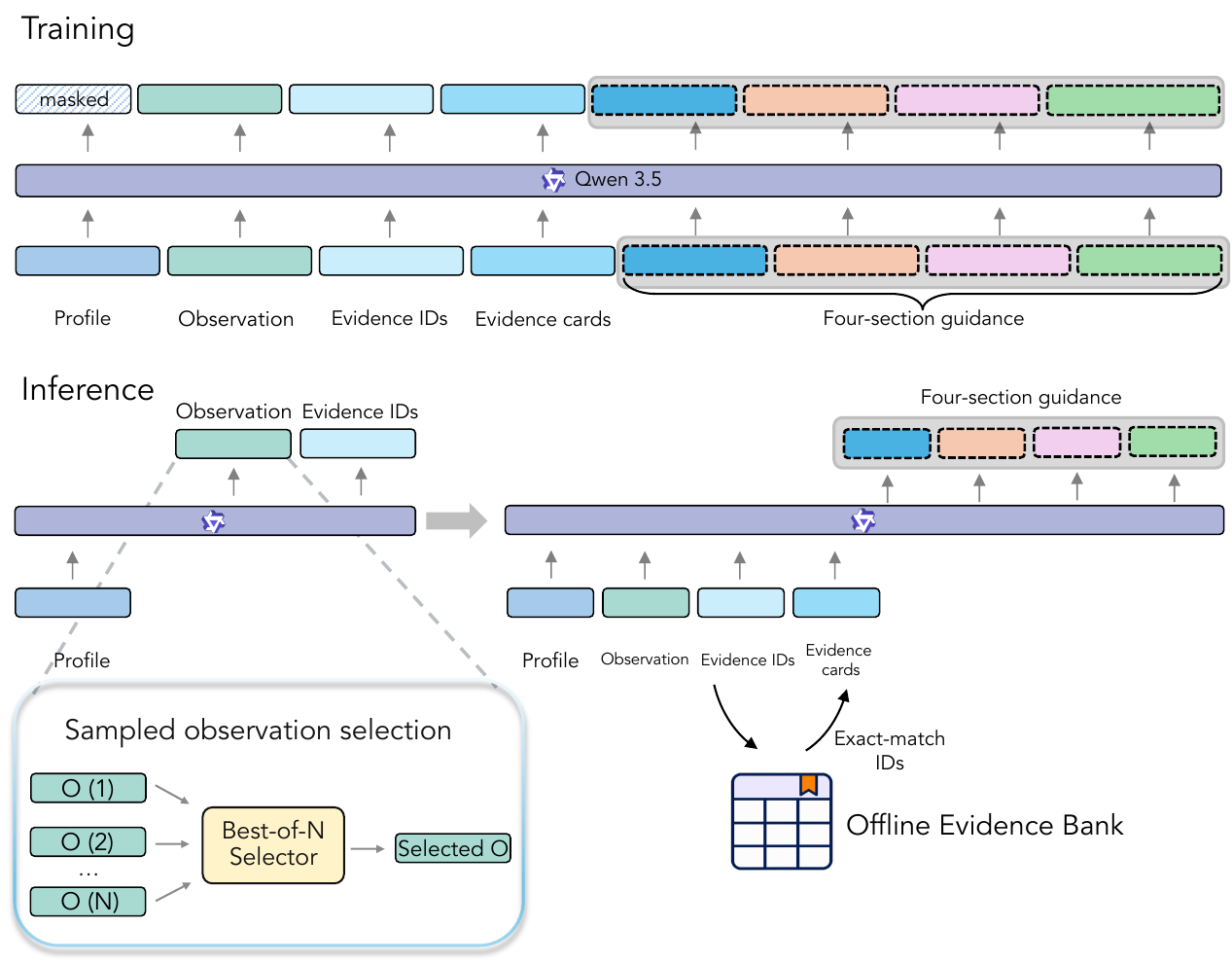}
\caption{Stage~2 training and two-pass inference with Best-of-$N$ selection and relevant evidence retrieval.}
\label{fig:stage2}
\end{figure}

Each training case follows the reasoning trajectory in a single autoregressive sequence:
\begin{equation}
[P_i;O_i^\star;E_i^\star;C_i^\star;R_i^\star],
~\text{where}~
C_i^\star=\mathcal B[E_i^\star].
\label{eq:stage2-sequence}
\end{equation}
The observation $O_i^\star$
captures salient multi-night sleep patterns before the evidence identifiers
$E_i^\star$ are selected, and the retrieved evidence cards $C_i^\star$
condition the generation of the final guidance $R_i^\star$. 
Unlike Guidance-only SFT, which supervises only the final guidance~$R_i^\star$ given $P_i$ and thus learns a direct input-to-answer mapping, our formulation supervises the 
reasoning trajectory.
Let
$
Z_i^\star=
[O_i^\star;E_i^\star;C_i^\star;R_i^\star]
$
denote the supervised completion 
given
$P_i$. We train the model using the autoregressive cross-entropy objective:
\begin{equation}
\mathcal L(\theta)
=
-\frac{1}{N_{\mathrm{tok}}}
\sum_{i,t}
\log p_{\theta}
\!\left(
Z_{i,t}^\star
\mid
P_i,Z_{i,<t}^\star
\right),
\label{eq:stage2-loss}
\end{equation}
where 
$N_{\mathrm{tok}}$ is the total number of target tokens across training cases,
$Z_{i,t}^\star$ is the $t^{\text{th}}$ token, and
$Z_{i,<t}^\star$ denotes its preceding tokens. 
During training, $C_i^\star$ remains in the target sequence so that the
model learns the complete reasoning trajectory. 
At inference, however, the model
generates the observation and evidence identifiers, while the corresponding
evidence cards are retrieved verbatim from the Evidence Bank (Figure~\ref{fig:stage2}).
This separation allows the model to learn which evidence
is relevant while preserving the exact retrieved
evidence content,
reducing the risk of unsupported evidence in the 
guidance generation.

Inference reuses the same model parameters $F_{\theta}(\cdot)$ in two passes. 
In the first pass, the~model generates
$\widehat O_i$ and $\widehat E_i$ from $P_i$ and stops at
\texttt{</EVIDENCE>}. Valid identifiers are resolved by: %
$\widehat C_i=\mathcal B[\widehat E_i]$.
In the second pass, the same model generates the final four-section guidance using
the retrieved evidence as part of the input:
\begin{equation}
\begin{aligned}
&\text{Pass 1:}\quad
(\widehat O_i,\widehat E_i)
=F_{\theta}(P_i), 
\widehat C_i =\mathcal B[\widehat E_i],\\
&\text{Pass 2:}~
\widehat R_i
=F_{\theta}
([P_i;\widehat O_i;\widehat E_i;\widehat C_i]).
\end{aligned}
\label{eq:stage2-inference}
\end{equation}

This two-pass procedure transfers the guidance trajectory to SLMs while
retrieving evidence deterministically, without rerunning the Stage~1
pipeline at deployment. 
Decoding and handling invalid identifiers 
are detailed in Appendix~\ref{app:details}.

\subsection{Best-of-$N$ Observation Selection}
\label{sec:assembly}

Because 
$\widehat O_i$ is the first generated component of the reasoning
trajectory, errors in $\widehat O_i$ can propagate to evidence selection and,
subsequently, to the final guidance. 
We therefore introduce a training-free
Best-of-$N$ strategy that samples multiple candidate observations and selects the best
one before evidence-identifier generation.

For case $i$, we sample an observation  $N{>}1$ times at a temperature of $0.8$, stopping each generation at \texttt{</OBSERVATION>}. Each candidate is
evaluated using two complementary signals: \emph{case-specific axis relevance}
and \emph{verdict-clause consensus}. The former depends on 
the 
individual's
longitudinal 
profile, 
whereas the latter can be compared directly across
candidates because verdict clauses follow fixed canonical forms.

\emph{Case-specific axis relevance.}
For candidate $n$, let $\mathcal M_n$ denote its metric--scope axis set, where
each axis encodes a sleep metric and its temporal or comparison scope, including
longitudinal changes. KNN over
profile features retrieves the
$K$ nearest training cases, whose axes are aggregated by similarity-weighted
voting to form the expected axis set $\widehat{\mathcal M}_i$. 
The 
$K$ was selected on training data for candidate discrimination.

\emph{Verdict-clause consensus.}
Let $\mathcal V_n$ denote the canonical verdict-clause set of candidate $n$.
Each clause encodes a predefined comparison between a case measurement and a
reference value specified in the Evidence Bank and maps to the corresponding
evidence cards. Agreement across candidates therefore reflects consensus on
which reference evidence applies to the case.

Candidate $n$ is scored as:
\begin{equation}
s_n
=
\frac{1}{2}
F_1\!\left(
\mathcal M_n,\widehat{\mathcal M}_i
\right)
+
\frac{1}{2}
\frac{1}{N-1}
\sum_{n'\neq n}
F_1\!\left(
\mathcal V_n,\mathcal V_{n'}
\right),
\label{eq:bon-score}
\end{equation}
where $F_1(\cdot)$ is cosine similarity. 
The first 
term
measures agreement with the KNN-derived 
axis expectation, and the second measures verdict-clause consensus across the
$N$ candidates.

The candidate with the highest $s_n$ is selected as $\widehat O_i$. Evidence
identifier generation, deterministic Evidence Bank lookup, and guidance
generation then proceed as described in Section~\ref{sec:stage2}. 
The selector requires
no 
training; 
see Appendix~\ref{app:details} for details.

\section{Experiments}
\label{sec:experiments}

\subsection{Setup}
\label{sec:setup}

All experiments use the participant-level split 
described in
Section~\ref{sec:data}, with
$1{,}200$ training cases and $300$ held-out test
cases. Our method uses Best-of-$N$ observation selection with $N=16$.

\paragraph{Baselines.}
We compare against both commercial general and medical LLMs, as well as open-source models.
For the former, 
we evaluate GPT-5.6-Sol, GPT Health, and OpenEvidence (Osler). These models receive the same profile packet as our method, together with the full text of all 22 evidence cards.
For open-source models, \emph{Guidance-only SFT} uses the same Qwen3.5 backbone with 0.8B, 2B, and 4B parameters, training cases, and
optimization schedule as our method, but supervises only input-guidance pairs ($P_i, R_i^\star$) without reasoning trajectories. 

\begin{table*}[!t]
\centering
\footnotesize
\setlength{\tabcolsep}{3pt}
\renewcommand{\arraystretch}{0.70}
\caption{\textbf{Comparison of ours and baselines.}
All rows use the same $300$ held-out test cases. Multi-seed results of trainable models report mean$\pm$sd. \frozenicon marks frozen prompting models; \finetunedicon marks fine-tuned models. LLM-as-a-Judge scores use a 1--5 Likert scale.}
\label{tab:main}

\begin{tabular}{lccccc}
\toprule
Method & Evidence F1 & Value F1 & ROUGE-L & BERTScore & LLM-as-a-Judge \\
\midrule

\frozenicon\ GPT-5.6-Sol
& 0.288
& 0.404
& 0.251
& 0.312
& 3.48 \\

\frozenicon\ GPT Health 
& 0.423
& 0.398
& 0.276
& 0.331
& 3.89 \\

\midrule

\frozenicon\ OpenEvidence
& 0.424
& 0.261
& 0.213
& 0.264
& 3.79 \\

\midrule

\finetunedicon\ Qwen3.5-4B, Guidance-only SFT
& \sdc{0.629}{0.016}
& \sdc{0.584}{0.008}
& \sdc{0.474}{0.005}
& \sdc{0.535}{0.005}
& 3.97 \\

\midrule

\rowcolor{hlblue}
\finetunedicon\ Ours (Qwen3.5-0.8B)
& \sdc{0.680}{0.010}
& \sdc{0.597}{0.005}
& \sdc{0.502}{0.003}
& \sdc{0.561}{0.002}
& 4.08 \\

\rowcolor{hlblue}
\finetunedicon\ Ours (Qwen3.5-2B)
& \sdc{0.707}{0.007}
& \sdc{0.609}{0.004}
& \sdc{0.510}{0.004}
& \sdc{0.567}{0.002}
& 4.14 \\

\rowcolor{hlblue}
\finetunedicon\ Ours (Qwen3.5-4B)
& \sdc{0.711}{0.006}
& \sdc{0.612}{0.003}
& \sdc{0.506}{0.001}
& \sdc{0.564}{0.001}
& 4.11 \\

\bottomrule
\end{tabular}

\end{table*}

\paragraph{Implementation.}
Stage~1 uses GPT-5.6-Sol with medium reasoning for the Profile, Evidence,
Composer, and Reflection Agents. Each agent uses a fresh context for each role invocation
to preserve role isolation.
Stage~2 fine-tunes Qwen3.5 models at $0.8$B, $2$B, and $4$B using the same
training configuration across model sizes.
Fine-tuned systems are trained with three random seeds and reported as
mean$\pm$sd, while prompting baselines are single runs.
For the paired comparison between greedy and Best-of-$N$ observation decoding
in Section~\ref{sec:exp-assembly}, we additionally report $95\%$ two-level
bootstrap confidence intervals over seeds and test cases (Appendix~\ref{app:ablation}).
Training hyperparameters are provided in
Appendix~\ref{app:details}.

\paragraph{Metrics.}
The main comparison reports five complementary metrics.
\emph{Evidence F1} compares the evidence recoverable from
a generated guidance with the verified Stage~1 evidence identifiers
$E_i^\star$. \emph{Value F1} measures recovery of the certified numeric
anchors $A_i^\star$. ROUGE-L \citep{lin2004rouge} and BERTScore
\citep{zhang2020bertscore} measure lexical and semantic similarity to
$R_i^\star$, respectively, but are not treated as evidence-grounding measures
\citep{maynez2020faithfulness,fabbri2021summeval}. \emph{LLM-as-a-Judge} based on Claude Opus 5 evaluates
generation quality using the Generation Quality Evaluation Rubric. All five metrics are evaluated on the full set of
$300$ held-out test cases. Human evaluation is described in Section~\ref{sec:human-eval}, with rubric details in Appendix~\ref{app:rubric}.

\subsection{Quantitative Evaluation}
\label{sec:exp-supervision}

As shown in Table~\ref{tab:main}, our method substantially outperforms commercial general and medical LLMs in personalized sleep guidance across Qwen3.5 with 0.8B, 2B, and 4B parameters. Using a 0.8B-parameter backbone, our model achieves an Evidence F1 of 0.680, compared with 0.288 for GPT-5.6-Sol and 0.423 for GPT Health. It also substantially outperforms OpenEvidence, which achieves an Evidence F1 of 0.424. In terms of LLM-as-a-Judge, our 0.8B model achieves a score of 4.08, compared with 3.48 for GPT-5.6-Sol, 3.89 for GPT Health, and 3.79 for OpenEvidence. Although GPT Health and OpenEvidence are designed for health-related applications, their evidence recovery remains below that of our method. These commercial general and medical LLMs generate guidance without the consistency checks and independent verification used in our framework to ensure alignment with the original wearable records and supporting evidence. These results suggest the effectiveness of our method and demonstrate the potential for SLMs to outperform general-purpose LLMs by constructing the evidence-grounded guidance and distilling the guidance reasoning trajectories.  

Our method also outperforms Guidance-only SFT, which is trained on the same data but learns only the final guidance without guidance reasoning trajectories. Notably, our 0.8B model outperforms the 4B Guidance-only SFT baseline across all five metrics, improving Evidence F1 from 0.629 to 0.680 and LLM-as-a-Judge from 3.97 to 4.08, with consistent gains in Value F1, ROUGE-L, and BERTScore. These results suggest that our framework makes effective use of the guidance information constructed during data generation, enabling guidance reasoning capability to be transferred to smaller models.

Additional analyses show consistent performance across age, sex, and record length, and robustness to missing days: with 30\% of days removed, our 0.8B model still achieves an Evidence F1 of 0.545, above all commercial baselines evaluated on complete records (Table~\ref{tab:missing}). In terms of inference cost, Best-of-16 selection adds only 6.7 seconds per case over greedy decoding (Appendix~\ref{app:ablation}).

\begin{table*}[t]
\centering
\footnotesize
\setlength{\tabcolsep}{5pt}
\renewcommand{\arraystretch}{1.08}
\caption{\textbf{Results of ablation studies.} All methods use Qwen3.5 with 0.8B parameters. Result values are mean$\pm$sd across three training seeds.}
\label{tab:ablation}
\begin{tabular}{lcccc}
\toprule
Method & Evidence F1 & Value F1 & ROUGE-L & BERTScore \\
\midrule
w/o Guidance Trajectory Distillation

& \sdc{0.536}{0.014}
& \sdc{0.556}{0.006}
& \sdc{0.447}{0.007}
& \sdc{0.511}{0.006} \\

w/o Best-of-$N$ selection

& \sdc{0.613}{0.019}
& \sdc{0.598}{0.005}
& \sdc{0.480}{0.009}
& \sdc{0.541}{0.008} \\

\rowcolor{hlblue}
Ours
& \sdc{\textbf{0.680}}{0.010}
& \sdc{0.597}{0.005}
& \sdc{\textbf{0.502}}{0.003}
& \sdc{\textbf{0.561}}{0.002} \\
\bottomrule
\end{tabular}
\end{table*}

\subsection{Human Evaluation}
\label{sec:human-eval}

\begin{figure}[t]
    \centering
    \includegraphics[width=\columnwidth]
    {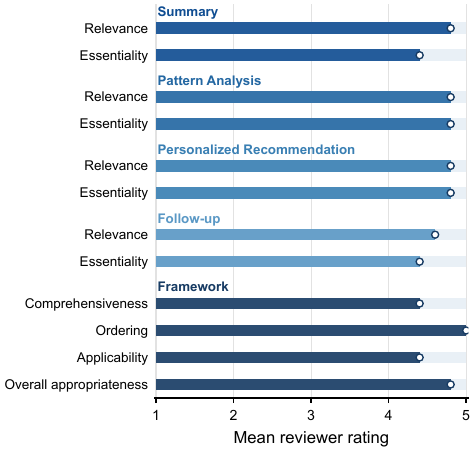}
    \caption{\textbf{Human evaluation of the Stage-1 guidance structure.}
    Mean item-wise ratings from five raters, each of whom reviewed ten
    distinct Stage~1 examples (Table~\ref{tab:stage1-rubric}).}
    \label{fig:stage1_ratings}
    \vspace{-20pt}
\end{figure}

\textbf{We first evaluated the four-section guidance structure produced in Stage~1.} Five raters (physicians and medical students) each reviewed ten distinct Stage~1 examples (50 in total) and rated the structure on a 12-item five-point rubric. The mean score was 4.67, with 97\% of ratings at 4 or 5 (Figure~\ref{fig:stage1_ratings}). These findings support the appropriateness of the proposed four-section structure for personalized sleep guidance.
Full rubric details are provided in Appendix~\ref{app:rubric-stage1}.

\begin{figure}[t]
    \centering
    \includegraphics[width=\columnwidth]
    {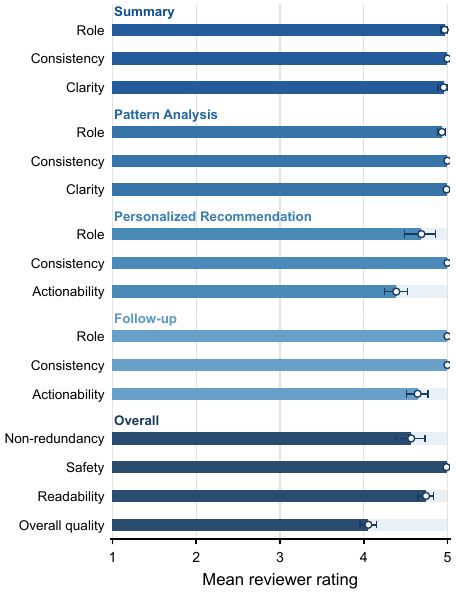}
    \caption{\textbf{Human evaluation of generated guidance.}
    Mean item-wise ratings with case-bootstrap 95\% confidence intervals
    across 90 evaluated cases.}
    \label{fig:human_ratings}
\end{figure}

\textbf{We then evaluated the guidance generated by our Stage~2 model.} Five experienced MD students rated 90 guidance examples generated by Qwen3.5-0.8B, with three independent ratings per example across 16 rubric items. Ratings were consistently high (\mbox{Figure~\ref{fig:human_ratings}}), including mean scores of 4.99 for safety, 4.74 for readability, and 4.06 for overall quality, with strong inter-rater agreement (ordinal Gwet's AC2 of 0.922; 95\% CI: 0.912--0.932). Full rubric results and pairwise agreement analyses are provided in
Appendix~\ref{app:rubric-stage2}, and more sleep guidance examples in Appendix~\ref{app:stage2-examples}.

\textbf{Finally, we compared our model with two baselines, with the wearable record shown to raters.} The evaluation above rated our model alone, and raters did not see the wearable record. Five raters (physicians and medical students) scored our 0.8B model, GPT Health, and Guidance-only SFT (4B, given all 22 evidence cards; Section~\ref{sec:exp-assembly}) on 30 test cases, with the three outputs shown in random order and each output rated by two raters. Raters also checked each output against the record, using two added items: \emph{data accuracy} and \emph{use of the data}. As shown in Table~\ref{tab:human-compare}, our model is rated highest on all four measures. Inter-rater agreement is high (Gwet's AC2 of 0.94); details are provided in Appendix~\ref{app:human-compare}.

\begin{table}[t]
\centering
\footnotesize
\setlength{\tabcolsep}{3pt}
\caption{\textbf{Comparative human evaluation with the wearable record available.} Values are per-case mean ratings (1--5) over 30 test cases, with two raters per output. Guidance-only SFT uses Qwen3.5-4B and is given all 22 evidence cards.}
\label{tab:human-compare}
\begin{tabular}{@{}lccc@{}}
\toprule
 & Ours & GPT & Guidance-only \\
Rubric item & (0.8B) & Health & SFT (4B) \\
\midrule
16 original items & \textbf{4.87} & 4.65 & 4.69 \\
Overall quality & \textbf{4.47} & 3.80 & 3.90 \\
Data accuracy & \textbf{4.98} & 4.08 & 4.67 \\
Use of the data & \textbf{4.88} & 4.43 & 4.28 \\
\bottomrule
\end{tabular}
\end{table}

\subsection{Ablation Study}
\label{sec:exp-assembly}

We conduct an ablation study on Qwen3.5-0.8B to examine the contributions of the main components of our method, with corresponding results for the 2B and 4B backbones provided in Appendix~\ref{app:ablation}. The results are shown in Table~\ref{tab:ablation}.

\emph{w/o Guidance Trajectory Distillation} only uses the final guidance without reasoning trajectories as the training target. Compared with this setting, our method achieves better performance across all metrics, increasing Evidence F1 from 0.536 to 0.680, Value F1 from 0.556 to 0.597, ROUGE-L from 0.447 to 0.502, and BERTScore from 0.511 to 0.561. The \emph{w/o Best-of-$N$ selection} setting retains the guidance trajectory as part of the supervised target and also outperforms the \emph{w/o Guidance Trajectory Distillation} setting across all metrics. These results demonstrate the benefit of incorporating the guidance trajectory into the supervised training target.

We further examine whether the improvement stems from access to the evidence itself by providing Guidance-only SFT with the full text of all 22 evidence cards during both training and inference. Its Evidence F1 changes little (0.536 vs.\ 0.556 at 0.8B; 0.629 vs.\ 0.620 at 4B) and remains below our 0.8B model (Appendix~\ref{app:ablation}). These results suggest that the gain comes from the guidance reasoning trajectory rather than from evidence access. Within the trajectory, both the evidence identifiers and the evidence-card text contribute to the gain (Table~\ref{tab:evidence-step} in Appendix~\ref{app:ablation}).

We study the effect of Best-of-$N$ observation selection to investigate how much selecting a suitable observation influences the generation performance. The \emph{w/o Best-of-$N$ selection} setting uses greedy observation decoding at inference, while our method applies Best-of-$N$ selection. Compared with greedy decoding, Best-of-$N$ selection increases Evidence F1 from 0.613 to 0.680, ROUGE-L from 0.480 to 0.502, and BERTScore from 0.541 to 0.561, while Value F1 remains comparable (0.598 vs.\ 0.597). These results show that the quality of the selected observation has a clear impact on 
performance, particularly on evidence alignment and guidance quality. We use $N=16$ in the main experiments. The effect of candidate-pool size
is reported in Table~\ref{tab:n-sweep} of Appendix~\ref{app:ablation}.

\begin{table}[t]
\centering
\scriptsize
\setlength{\tabcolsep}{3pt}
\renewcommand{\arraystretch}{1.08}
\caption{\textbf{Robustness of our 0.8B model to missing days.}
Values are mean$\pm$sd over three seeds.}
\label{tab:missing}
\begin{tabular}{@{}lccc@{}}
\toprule
Days removed & Evidence F1 & ROUGE-L & BERTScore \\
\midrule
None & \sdc{0.680}{0.010} & \sdc{0.502}{0.003} & \sdc{0.561}{0.002} \\
10\% & \sdc{0.601}{0.025} & \sdc{0.460}{0.013} & \sdc{0.524}{0.011} \\
20\% & \sdc{0.553}{0.017} & \sdc{0.442}{0.007} & \sdc{0.507}{0.008} \\
30\% & \sdc{0.545}{0.008} & \sdc{0.431}{0.003} & \sdc{0.497}{0.003} \\
\bottomrule
\end{tabular}
\end{table}

\section{Conclusion}

We presented a two-stage framework for personalized sleep guidance from longitudinal wearable data. 
A multi-agent pipeline constructs an evidence-grounded dataset, 
while guidance trajectory distillation transfers the reasoning process into SLMs. Training-free Best-of-$N$ observation selection further improves 
performance. 
Our method with the 0.8B model size outperformed commercial LLMs (including GPT-5.6-Sol, GPT Health, and OpenEvidence) and the 4B baseline.
Rubric-based human evaluation further supported 
both 
the proposed four-section 
guidance 
structure and the
generated guidance quality.

Our study is limited to Fitbit-derived records from the All of Us Research Program with 14--60 consecutive days of complete measurements, which may limit generalizability. The Evidence Bank supports behavioral guidance rather than diagnosis or treatment. Future work should integrate clinical information with wearable data and expand the 
Evidence Bank
to support more comprehensive personalized sleep guidance.

\acks{G.C.W. is supported in part by the National Cancer Institute of the National Institutes of Health under Award Number R00CA286966 and the Department of Defense under award number HT9425261E304.}

\bibliography{ref}

\clearpage
\appendix

\section{Additional Details}
\label{app:details}

\paragraph{Wearable variables and profile metrics.}
\label{app:data-metrics}
Each daily record carries sleep duration, time in bed, sleep timing, sleep
efficiency, device-estimated wake after sleep onset (WASO), sleep-stage minutes,
and activity measures. The profile packet $P_i$ preserves these measurements
together with age, sex, dates, window boundaries, and derived basic statistics used to
summarize the record.

\paragraph{Participant selection.}
\label{app:cohort}
Participants in the All of Us Research Program were eligible if they were aged
18 years or older, had female or male sex recorded, and had at least 14
consecutive days with complete Fitbit sleep and activity data. For each
eligible participant, we randomly selected one period of 14--60 consecutive
complete days as the record. We then randomly sampled 1,500 participants, 750
female and 750 male, such that the distribution of record lengths matches that
of all eligible participants. Table~\ref{tab:cohort} summarizes the resulting sample.

\begin{table}[!htbp]
\centering
\footnotesize
\setlength{\tabcolsep}{4pt}
\caption{\textbf{Characteristics of the 1,500 participants.}}
\label{tab:cohort}
\begin{tabular}{@{}lc@{}}
\toprule
Characteristic & Value \\
\midrule
Sex, female / male & 750 / 750 \\
Age (years), median (IQR), range & 58 (41--69), 18--90 \\
Days per record, median (IQR) & 20 (16--29) \\
Sleep duration (h), mean $\pm$ SD & $6.6 \pm 0.9$ \\
\bottomrule
\end{tabular}
\end{table}

\paragraph{Evidence Bank construction and validation.}
The Evidence Bank contains 22 evidence cards, and each evidence identifier
refers to one of them. The cards were built around the sleep dimensions that a
consumer wearable measures: duration, continuity, timing and regularity,
latency, efficiency, stage estimates, and activity. For each dimension, we
collected authoritative sources, including clinical guidelines and consensus
statements (e.g., the American Academy of Sleep Medicine and the National Sleep
Foundation), systematic
reviews and meta-analyses, original studies, government health guidance, and
device documentation. In total, the cards are derived from 46 sources, each
recorded with a resolvable DOI or URL. Each card states its content (a reference value, an interpretation,
or a recommendation), the condition under which it applies, and the claims it
must not make. For example, the sleep-duration card for adults aged 18--64 is
derived from the National Sleep Foundation sleep duration recommendations: it
gives a reference value of at least 7 hours, applies when a person aged 18--64
repeatedly sleeps clearly less than 7 hours, and must not be applied to a
single short night or to adults aged 65 or older. Together, the cards cover
every sleep and activity measure in the profile packet.
Table~\ref{tab:cards} lists the evidence statement of each card.

During Stage~1, the Evidence Agent can select
only existing identifiers, retrieved card text is inserted verbatim, and
deterministic checks reject unsupported conditions and numeric anchors. 
If an evidence check fails, evidence selection and composition are restarted;
if a check against the participant's data fails, the draft is returned to the
Composer Agent and the selected evidence is retained.

\begin{table*}[p]
\centering
\scriptsize
\setlength{\tabcolsep}{4pt}
\renewcommand{\arraystretch}{1.05}
\caption{\textbf{The 22 evidence cards in the Evidence Bank.} Each row gives
the card and its evidence statement (abridged from the card text). Full
bibliographic details of the 46 sources from which the cards are derived are
available in the code repository.}
\label{tab:cards}
\begin{tabular}{@{}r>{\raggedright\arraybackslash}p{0.30\textwidth}>{\raggedright\arraybackslash}p{0.62\textwidth}@{}}
\toprule
\# & Evidence card & Evidence statement \\
\midrule
1 & Sleep duration reference for adults aged 18 to 64 & Adults generally need 7 or more hours of sleep, although individual needs vary. \\
2 & Sleep duration reference for adults aged 65 and older & Older adults generally need 7 or more hours of sleep, although individual needs vary. \\
3 & Sleep efficiency quality reference & Spending at least 85\% of time in bed asleep across repeated nights is a general sign of good sleep quality. \\
4 & Time awake after sleep begins in adults & For adults, 20 minutes or less awake after first falling asleep is a general sign of good sleep quality; repeatedly longer times can indicate interrupted sleep. \\
5 & Time awake after sleep begins in older adults & For older adults, 30 minutes or less awake after first falling asleep is a general sign of good sleep quality; repeatedly longer times can indicate interrupted sleep. \\
6 & Sleep-latency quality reference for adults & Falling asleep within 30 minutes is a general sign of good sleep quality for adults; repeatedly longer times can indicate difficulty settling to sleep. \\
7 & Longer wake-episode quality reference for adults aged 18 to 64 & For adults, no more than one wake episode lasting over five minutes per night is a general sign of good sleep quality. \\
8 & Longer wake-episode quality reference for adults aged 65 and older & For older adults, no more than two wake episodes lasting over five minutes per night is a general sign of good sleep quality. \\
9 & Time in bed and sleep duration describe different parts of the night & Time in bed includes time before falling asleep, wakefulness during the night, and time after the final wake, so it is not the same measure as sleep duration. \\
10 & Allow more time for sleep & When sleep is repeatedly short and the time available for sleep is also limited, making more time available can increase sleep duration. \\
11 & Make sleep timing more consistent & Regular sleep timing is one dimension of sleep health, but no universal bedtime, wake time, or acceptable variation applies to every person. \\
12 & Preserve recovery sleep when weekday sleep is short & When weekday sleep is repeatedly short, longer weekend sleep may partly reflect recovery, so reducing the timing difference should not take priority over preserving sleep duration. \\
13 & Maintain a stable adequate sleep pattern & Adequate duration, appropriate timing, regularity, and good quality are complementary dimensions of healthy sleep; a stable adequate pattern can be maintained when no repeated concern outweighs it. \\
14 & Caffeine close to sleep may disrupt sleep & Caffeine near bedtime can shorten sleep, reduce sleep efficiency, and increase time awake after sleep begins. \\
15 & Late light or screen use may delay sleep & Bright light late in the evening can signal wakefulness and may contribute to later sleep timing. \\
16 & A quiet wind-down may support sleep & A quiet period before bed may make it easier to settle for sleep when the period before sleep is busy or stimulating. \\
17 & Regular daytime movement may support sleep efficiency & Regular daytime movement can support sleep quality, although a within-person association between activity and sleep efficiency does not prove causation. \\
18 & Sleep-stage estimates are descriptive, not action targets & Deep and REM sleep estimates vary from night to night and may differ from laboratory sleep staging, so they should not be treated as precise personal targets. \\
19 & Sleep timing alone does not confirm circadian phase & Sleep start, wake time, and the midpoint of sleep describe a timing pattern, but clock times alone cannot confirm the timing of the internal body clock or a circadian disorder. \\
20 & Unwanted sleep timing may need clinical review & A stable late or early schedule is not automatically a disorder; a persistent timing shift warrants professional evaluation only when it is unwanted and conflicts with daily responsibilities. \\
21 & Persistent short sleep needs clinical context & When short sleep continues despite enough time in bed, sleep history and daytime symptoms help a clinician decide whether schedule demands, another health issue, or a sleep disorder needs attention. \\
22 & Persistent sleep difficulty needs clinical context & A repeated sleep pattern is more useful to discuss with a clinician when the person also has ongoing trouble sleeping, unusual daytime sleepiness, or trouble concentrating; the sleep record alone cannot establish a disorder. \\
\bottomrule
\end{tabular}
\end{table*}

\paragraph{Prompts for the commercial baselines.}
GPT-5.6-Sol, GPT Health, and OpenEvidence receive one prompt per case, without
few-shot examples, tools, or prompt optimization. The prompt consists of
(i) the instructions of the Composer Agent for writing the four-section
guidance (Appendix~\ref{app:composer-prompt}); (ii) the profile packet $P_i$,
identical to the input of our model; (iii) the full text of all 22 evidence
cards; and (iv) the output format, which asks for an observation, the identifiers of the evidence cards used, and the four sections. Unlike the Composer Agent, which receives
the cards already selected by the Evidence Agent and whose output is reviewed
by the Reflection Agent, the baselines choose among all 22 cards themselves
and write the guidance in a single response.

\paragraph{Stage-2 decoding and selection constants.}
The first decoding pass begins with \texttt{<OBSERVATION>} and stops at
\texttt{</EVIDENCE>}. Identifiers are matched exactly against the bank;
duplicates are removed and invalid identifiers count as errors. The decoder
then inserts the retrieved cards with \texttt{<CARDS>}, appends
\texttt{<SECTIONS>}, and resumes generation with the same model. For Best-of-$N$,
packet similarity is cosine similarity over a standardized deterministic
feature vector. The $K=12$ nearest training cases vote with similarity-softmax
weights at temperature $8.0$, and an axis enters the expected set when its
weighted vote share reaches $0.4$.

\paragraph{Training hyperparameters.}
Stage~2 uses completion-only cross-entropy, AdamW, bfloat16 precision, a learning
rate of $2\times10^{-5}$ with cosine decay and $5\%$ warmup, weight decay $0.1$,
gradient clipping at $1.0$, an effective batch size of 16, five epochs, and a
maximum sequence length of 32,768 tokens. The three model sizes share this
configuration.

\section{Additional Results}
\label{app:ablation}

\begin{table*}[t]
\centering
\footnotesize
\setlength{\tabcolsep}{3.5pt}
\renewcommand{\arraystretch}{1.08}
\caption{\textbf{Ablation across Qwen3.5 model sizes.}
Values are mean$\pm$sd over three seeds.}
\label{tab:ablation-full}

\begin{tabular}{llcccc}
\toprule
Model & Method & Evidence F1 & Value F1 & ROUGE-L & BERTScore \\
\midrule

0.8B
& w/o Guidance Trajectory Distillation
& \sdc{0.536}{0.014}
& \sdc{0.556}{0.006}
& \sdc{0.447}{0.007}
& \sdc{0.511}{0.006} \\

& w/o Best-of-$N$ selection
& \sdc{0.613}{0.019}
& \sdc{0.598}{0.005}
& \sdc{0.480}{0.009}
& \sdc{0.541}{0.008} \\

\rowcolor{hlblue}
& Ours
& \sdc{0.680}{0.010}
& \sdc{0.597}{0.005}
& \sdc{0.502}{0.003}
& \sdc{0.561}{0.002} \\

\addlinespace[3pt]

2B
& w/o Guidance Trajectory Distillation
& \sdc{0.603}{0.013}
& \sdc{0.565}{0.010}
& \sdc{0.472}{0.009}
& \sdc{0.533}{0.008} \\

& w/o Best-of-$N$ selection
& \sdc{0.653}{0.012}
& \sdc{0.606}{0.005}
& \sdc{0.494}{0.004}
& \sdc{0.553}{0.005} \\

\rowcolor{hlblue}
& Ours
& \sdc{0.707}{0.007}
& \sdc{0.609}{0.004}
& \sdc{0.510}{0.004}
& \sdc{0.567}{0.002} \\

\addlinespace[3pt]

4B
& w/o Guidance Trajectory Distillation
& \sdc{0.629}{0.016}
& \sdc{0.584}{0.008}
& \sdc{0.474}{0.005}
& \sdc{0.535}{0.005} \\

& w/o Best-of-$N$ selection
& \sdc{0.693}{0.001}
& \sdc{0.614}{0.002}
& \sdc{0.502}{0.003}
& \sdc{0.561}{0.002} \\

\rowcolor{hlblue}
& Ours
& \sdc{0.711}{0.006}
& \sdc{0.612}{0.003}
& \sdc{0.506}{0.001}
& \sdc{0.564}{0.001} \\

\bottomrule
\end{tabular}
\end{table*}

Table~\ref{tab:ablation-full} reports the same ablation as
Table~\ref{tab:ablation} across all three Qwen3.5 model sizes.
Within each size, the greedy and Best-of-$16$ variants use the same
three checkpoints and differ only in observation generation and selection.

Best-of-$16$ improves Evidence F1 over greedy decoding at $0.8$B and
$2$B, with confidence intervals that exclude zero. The improvement is
smaller at $4$B, where the confidence interval includes zero.

\paragraph{Effect of candidate-pool size.}
Table~\ref{tab:n-sweep} varies the number of candidate observations on
Qwen3.5-0.8B. Evidence F1 increases from
$0.591$ at $N=2$ to $0.680$ at $N=16$ and $0.692$ at $N=32$, while
the remaining metrics change by at most $0.007$ beyond $N{=}16$. The gains
beyond $N{=}16$ are therefore small.

\begin{table}[t]
\centering
\scriptsize
\setlength{\tabcolsep}{2pt}
\renewcommand{\arraystretch}{1.08}
\caption{\textbf{Effect of candidate-pool size on Qwen3.5-0.8B.}
Greedy and Best-of-$16$ report the three-seed means of Table~\ref{tab:ablation}; the other rows use one checkpoint.}
\label{tab:n-sweep}

\begin{tabular}{lcccc}
\toprule
Decoding & Ev. F1 & Val. F1 & R-L & BERTScore \\
\midrule
Greedy
& 0.613
& 0.598
& 0.480
& 0.541 \\

Best-of-$2$
& 0.591
& 0.576
& 0.469
& 0.531 \\

Best-of-$4$
& 0.629
& 0.587
& 0.483
& 0.542 \\

Best-of-$8$
& 0.681
& 0.592
& 0.499
& 0.560 \\

Best-of-$16$
& 0.680
& 0.597
& 0.502
& 0.561 \\

Best-of-$32$
& 0.692
& 0.604
& 0.504
& 0.564 \\
\bottomrule
\end{tabular}
\end{table}

\paragraph{Effect of evidence access.}
Table~\ref{tab:all-cards} examines whether access to the evidence alone
explains the improvement. We train Guidance-only SFT with the same training configuration as
our model and provide the full text of all 22 evidence cards during both
training and inference. At $0.8$B, Evidence F1 changes little ($0.536$ vs.\ $0.556$) and remains
$0.12$ below our model. At $4$B, Evidence F1 is unchanged
within one standard deviation ($0.629$ vs.\ $0.620$), and this larger model
remains below our $0.8$B model on all four metrics.

\begin{table*}[t]
\centering
\footnotesize
\setlength{\tabcolsep}{3.5pt}
\renewcommand{\arraystretch}{1.08}
\caption{\textbf{Guidance-only SFT with access to all 22 evidence cards.}
Values are mean$\pm$sd over three seeds.}
\label{tab:all-cards}
\begin{tabular}{lcccc}
\toprule
Method & Evidence F1 & Value F1 & ROUGE-L & BERTScore \\
\midrule
Guidance-only SFT (0.8B)
& \sdc{0.536}{0.014} & \sdc{0.556}{0.006} & \sdc{0.447}{0.007} & \sdc{0.511}{0.006} \\
Guidance-only SFT + all 22 cards (0.8B)
& \sdc{0.556}{0.010} & \sdc{0.559}{0.003} & \sdc{0.457}{0.004} & \sdc{0.521}{0.003} \\
Guidance-only SFT (4B)
& \sdc{0.629}{0.016} & \sdc{0.584}{0.008} & \sdc{0.474}{0.005} & \sdc{0.535}{0.005} \\
Guidance-only SFT + all 22 cards (4B)
& \sdc{0.620}{0.003} & \sdc{0.589}{0.012} & \sdc{0.475}{0.005} & \sdc{0.535}{0.004} \\
\rowcolor{hlblue}
Ours (0.8B)
& \sdc{0.680}{0.010} & \sdc{0.597}{0.005} & \sdc{0.502}{0.003} & \sdc{0.561}{0.002} \\
\bottomrule
\end{tabular}
\end{table*}

\paragraph{Ablation of the evidence step.}
Table~\ref{tab:evidence-step} compares variants of the reasoning trajectory
that differ only in how much of the evidence step they contain. Adding the
evidence identifiers to the observation raises Evidence F1 from $0.620$ to
$0.642$, and adding the evidence-card text raises it further to $0.680$.
Whether the card text is generated by the model or retrieved from the Evidence
Bank does not change the scores ($0.679$ vs.\ $0.680$). However, the generated
text differs from the card text in the bank in about 3 of 300 cases, whereas
retrieval inserts the verified card text in every case.

\begin{table*}[t]
\centering
\footnotesize
\setlength{\tabcolsep}{3.5pt}
\renewcommand{\arraystretch}{1.08}
\caption{\textbf{Ablation of the evidence step on Qwen3.5-0.8B.}
Values are mean$\pm$sd over three seeds.}
\label{tab:evidence-step}
\begin{tabular}{lcccc}
\toprule
Trajectory content & Evidence F1 & Value F1 & ROUGE-L & BERTScore \\
\midrule
Observation only
& \sdc{0.620}{0.029} & \sdc{0.587}{0.008} & \sdc{0.479}{0.010} & \sdc{0.541}{0.008} \\
Observation + identifiers
& \sdc{0.642}{0.018} & \sdc{0.605}{0.007} & \sdc{0.491}{0.005} & \sdc{0.549}{0.006} \\
Observation + identifiers + generated card text
& \sdc{0.679}{0.010} & \sdc{0.597}{0.006} & \sdc{0.502}{0.002} & \sdc{0.561}{0.002} \\
\rowcolor{hlblue}
Observation + identifiers + retrieved card text (ours)
& \sdc{0.680}{0.010} & \sdc{0.597}{0.005} & \sdc{0.502}{0.003} & \sdc{0.561}{0.002} \\
\bottomrule
\end{tabular}
\end{table*}

\paragraph{Subgroup analysis.}
Table~\ref{tab:subgroups} reports the performance of our $0.8$B model by age,
sex, and record length on the 300 held-out test cases, which span ages 18--84
(median 58) and records of 14--60 days. No difference between groups is
significant after correction for the seven pairwise comparisons (three between age groups, one between sexes, and three between record lengths).

\begin{table*}[t]
\centering
\footnotesize
\setlength{\tabcolsep}{3.5pt}
\renewcommand{\arraystretch}{1.08}
\caption{\textbf{Subgroup results of our 0.8B model on the held-out test set.}
Values are mean$\pm$sd over three seeds.}
\label{tab:subgroups}
\begin{tabular}{lccccc}
\toprule
Group & $n$ & Evidence F1 & Value F1 & ROUGE-L & BERTScore \\
\midrule
Age 18--39 & 65 & \sdc{0.614}{0.017} & \sdc{0.599}{0.012} & \sdc{0.485}{0.006} & \sdc{0.540}{0.005} \\
Age 40--64 & 131 & \sdc{0.661}{0.021} & \sdc{0.584}{0.008} & \sdc{0.491}{0.004} & \sdc{0.554}{0.005} \\
Age 65+ & 104 & \sdc{0.744}{0.026} & \sdc{0.612}{0.005} & \sdc{0.528}{0.005} & \sdc{0.582}{0.005} \\
\addlinespace[2pt]
Female & 161 & \sdc{0.667}{0.006} & \sdc{0.592}{0.007} & \sdc{0.494}{0.001} & \sdc{0.554}{0.000} \\
Male & 139 & \sdc{0.694}{0.023} & \sdc{0.602}{0.005} & \sdc{0.511}{0.004} & \sdc{0.568}{0.004} \\
\addlinespace[2pt]
14--20 days & 154 & \sdc{0.676}{0.013} & \sdc{0.609}{0.004} & \sdc{0.500}{0.005} & \sdc{0.560}{0.004} \\
21--40 days & 114 & \sdc{0.696}{0.013} & \sdc{0.586}{0.004} & \sdc{0.509}{0.007} & \sdc{0.566}{0.007} \\
41--60 days & 32 & \sdc{0.639}{0.063} & \sdc{0.578}{0.022} & \sdc{0.487}{0.021} & \sdc{0.545}{0.019} \\
\addlinespace[2pt]
All & 300 & \sdc{0.680}{0.010} & \sdc{0.597}{0.005} & \sdc{0.502}{0.003} & \sdc{0.561}{0.002} \\
\bottomrule
\end{tabular}
\end{table*}

\paragraph{Robustness to missing days.}
To simulate incomplete wearable logs, we randomly removed $10\%$, $20\%$, or $30\%$ of
the days from each test record, recomputed the profile packet from the
remaining days, and ran our full method without retraining
(Table~\ref{tab:missing}). With $30\%$ of days removed, Evidence F1 is $0.545$,
above all commercial baselines evaluated on complete records
($0.288$--$0.424$; Table~\ref{tab:main}). The guidance for every record still
contains the four sections. Value F1 is not reported because the reference was
written for the complete record, so the numbers it contains no longer match the
record once days are removed.

\paragraph{Inference time.}
Table~\ref{tab:latency} reports the inference time of Qwen3.5-0.8B on one
RTX PRO 6000 GPU. Best-of-$16$ adds $6.7$ seconds per case over greedy
decoding, and Best-of-$32$ adds another $9.3$ seconds for a further gain of only
$0.012$ in Evidence F1 (Table~\ref{tab:n-sweep}).

\begin{table}[H]
\centering
\scriptsize
\setlength{\tabcolsep}{6pt}
\renewcommand{\arraystretch}{1.08}
\caption{\textbf{Inference time of Qwen3.5-0.8B.}}
\label{tab:latency}
\begin{tabular}{lc}
\toprule
Decoding & Seconds per case \\
\midrule
Greedy & 10.7 \\
Best-of-$16$ & 17.4 \\
Best-of-$32$ & 26.7 \\
\bottomrule
\end{tabular}
\end{table}

\paragraph{Error analysis of GPT-5.6-Sol.}
We compared the evidence cited by GPT-5.6-Sol with the verified Stage~1
evidence on the test set and observed three types of errors.
\emph{(i) Excessive evidence.} GPT-5.6-Sol cites $4.7$ evidence cards per case,
compared with $2.0$ in the references, and cites more than the reference in
$99\%$ of cases.
\emph{(ii) Measure-level rather than pattern-level interpretation.} Whenever a
measure falls outside its reference value, GPT-5.6-Sol cites the evidence for
that measure, but it misses evidence whose relevance depends on the overall
pattern. For example, it never cites the evidence that supports keeping an
already adequate and stable sleep pattern unchanged.
\emph{(iii) Correct but excessive numbers.} $99\%$ of the numbers it states
appear in the participant's record, but it states about twice as many as the
references.
As an example, for a 69-year-old woman with 33 days of data, GPT-5.6-Sol cited
five evidence cards and recommended allowing more time for sleep, whereas the
reference cited three. The record shows that she already spends about 70
minutes more in bed than asleep, mostly awake after falling asleep, so her
short sleep reflects interrupted sleep rather than limited time in bed, and the
recommendation targets the wrong problem.

\section{Rubric Materials}
\label{app:rubric}

This appendix separates the administered rubric specifications from the
aggregate findings in Section~\ref{sec:human-eval}. For each stage, we report
the rated statements, fully labeled guidance options, and item-level scores.

\subsection{Stage-1 Framework Validation Rubric}
\label{app:rubric-stage1}

Each of the five raters read one uninterrupted batch of ten distinct
evidence-grounded examples before rating the framework once. The instrument contained eight section-level
items, relevance and essentiality for each of Summary, Pattern Analysis,
Personalized Recommendation, and Follow-up, followed by four framework-level
items. Table~\ref{tab:stage1-scales} gives the administered five-point scales,
and Table~\ref{tab:stage1-rubric} reports the ratings.

\begin{table}[H]
\centering
\scriptsize
\setlength{\tabcolsep}{2.6pt}
\renewcommand{\arraystretch}{1.06}

\caption{\textbf{Item-level Stage-1 framework ratings.}
Mean is the average rating over the five raters; I-CVI (item-level content validity index) is the proportion of raters assigning a score of 4 or 5.}
\label{tab:stage1-rubric}
\begin{tabular}{@{}>{\raggedright\arraybackslash}p{0.62\columnwidth}cc@{}}
\toprule
Item & Mean & I-CVI \\
\midrule
Summary relevance & 4.80 & 1.00 \\
Summary essentiality & 4.40 & 1.00 \\
Pattern Analysis relevance & 4.80 & 1.00 \\
Pattern Analysis essentiality & 4.80 & 1.00 \\
Personalized Recommendation relevance & 4.80 & 1.00 \\
Personalized Recommendation essentiality & 4.80 & 1.00 \\
Follow-up relevance & 4.60 & 1.00 \\
Follow-up essentiality & 4.40 & 0.80 \\
Comprehensiveness & 4.40 & 0.80 \\
Ordering & 5.00 & 1.00 \\
Applicability & 4.40 & 1.00 \\
Overall appropriateness & 4.80 & 1.00 \\
\bottomrule
\end{tabular}
\end{table}
\begin{table*}[htbp]
\centering
\scriptsize
\setlength{\tabcolsep}{4pt}
\renewcommand{\arraystretch}{1.12}
\caption{Stage-1 Framework Validation Rubric. Relevance and essentiality were
rated separately for each guidance section; the remaining constructs were
rated once for the framework as a whole.}
\label{tab:stage1-scales}
\begin{tabular}{@{}>{\raggedright\arraybackslash}p{0.15\textwidth}
>{\raggedright\arraybackslash}p{0.25\textwidth}
>{\raggedright\arraybackslash}p{0.53\textwidth}@{}}
\toprule
Construct & Rated statement & Guidance options \\
\midrule
Relevance & The content of this section is relevant to a sleep recommendation. &
\textbf{1}: Not relevant; it does not belong in a sleep recommendation. \textbf{2}: Slightly relevant; only a small part belongs. \textbf{3}: Moderately relevant; about half belongs. \textbf{4}: Relevant; most belongs. \textbf{5}: Highly relevant; all of it belongs. \\
Essentiality & This section is essential; the recommendation would be incomplete without it. &
\textbf{1}: Not necessary; it should be removed. \textbf{2}: Useful, but the recommendation works without it. \textbf{3}: Essential in some cases, but not all. \textbf{4}: Essential in most cases. \textbf{5}: Always essential; the recommendation would be incomplete without it. \\
Comprehensiveness & Together, the four sections cover the expected content. &
\textbf{1}: Most important content is missing. \textbf{2}: Several important elements are missing. \textbf{3}: One important element is missing. \textbf{4}: Only a minor element is missing. \textbf{5}: Nothing is missing. \\
Ordering & The sequence of the four sections follows a logical order. &
\textbf{1}: The order is illogical and would confuse the reader. \textbf{2}: At least one section is clearly misplaced. \textbf{3}: The order works, but another order would work equally well. \textbf{4}: The order is logical, with small room for improvement. \textbf{5}: The order is fully logical for the reader. \\
Applicability & The framework can be applied consistently across cases. &
\textbf{1}: It does not fit most cases. \textbf{2}: It fits some cases but breaks down in many others. \textbf{3}: It fits about half the cases well. \textbf{4}: It fits most cases, with occasional strain. \textbf{5}: It fits every case shown without strain. \\
Overall appropriateness & The four-section framework is appropriate as a standard structure. &
\textbf{1}: Not appropriate; redesign the framework. \textbf{2}: Weak; an important component needs to change. \textbf{3}: Acceptable, but another framework could work equally well. \textbf{4}: Appropriate, with only minor changes needed. \textbf{5}: Fully appropriate; keep it as is. \\
\bottomrule
\end{tabular}
\end{table*}

\paragraph{LLM-based assessment of Stage-1 examples.}
We additionally used Claude Opus 5 to assess the 50 Stage~1 examples reviewed
in the framework validation. For each case, the judge received the
participant's wearable record, all 22 evidence cards, the selected cards, and
the guidance, and rated two criteria on a 1--5 scale. \emph{Scientific
correctness} rates the accuracy of the scientific claims in the selected
evidence and the recommendations; the mean score was $4.74$, with $98\%$ of
cases at 4 or 5. \emph{Applicability to the participant} rates how well the selected evidence and
the recommendations suit the participant given the wearable record; the mean
score was $4.34$, with $84\%$ of cases at 4 or 5.

\subsection{Stage-2 Generation Quality Evaluation Rubric}
\label{app:rubric-stage2}

The Stage-2 instrument rated each generated four-section guidance on 16 items.
Twelve items assessed individual sections, and four assessed the guidance as a
whole. Table~\ref{tab:stage2-items} maps the items to their rated statements;
Table~\ref{tab:stage2-section-scales} gives the full score anchors. Table~\ref{tab:stage2-rubric-full} reports the
item-level human ratings.

\begin{table*}[htbp]
\centering
\scriptsize
\setlength{\tabcolsep}{4pt}
\renewcommand{\arraystretch}{1.08}
\caption{Stage-2 rubric item map. Role and consistency were rated for every
section; clarity was used for Summary and Pattern Analysis, and actionability
for Personalized Recommendation and Follow-up.}
\label{tab:stage2-items}
\begin{tabular}{@{}>{\raggedright\arraybackslash}p{0.045\textwidth}
>{\raggedright\arraybackslash}p{0.17\textwidth}
>{\raggedright\arraybackslash}p{0.13\textwidth}
>{\raggedright\arraybackslash}p{0.58\textwidth}@{}}
\toprule
Item & Scope & Construct & Rated statement \\
\midrule
1 & Summary & Role (A) & The section stays within its assigned role: summarize observed sleep data, without advice, causes, or clinical framing. \\
2 & Summary & Consistency (B) & The section does not contradict itself or the other sections. \\
3 & Summary & Clarity (C) & The section clearly presents all important sleep data needed. \\
4 & Pattern Analysis & Role (A) & The section interprets patterns as possibilities, without presenting unmeasured factors as facts or making clinical claims. \\
5 & Pattern Analysis & Consistency (B) & The section does not contradict itself or the other sections. \\
6 & Pattern Analysis & Clarity (C) & The section clearly presents all important sleep data needed. \\
7 & Personalized Recommendation & Role (A) & The section gives a few low-risk, reversible changes with reasons and a way to track them, or states that no change is needed; it gives no diagnosis, medication, supplement, or treatment content. \\
8 & Personalized Recommendation & Consistency (B) & The section does not contradict itself or the other sections. \\
9 & Personalized Recommendation & Actionability (D) & The reader knows what to do and how to track it. \\
10 & Follow-up & Role (A) & The section states when and how to reassess and when to talk to a clinician, without diagnostic claims, urgency, or new content. \\
11 & Follow-up & Consistency (B) & The section does not contradict itself or the other sections. \\
12 & Follow-up & Actionability (D) & The reader knows what to do and how to track it. \\
13 & Overall & Non-redundancy (E) & The four sections do not repeat each other. \\
14 & Overall & Safety (F) & The output contains nothing out of scope and nothing that could cause harm if followed. \\
15 & Overall & Readability (G) & The output is easy to read, with correct grammar, spelling, and formatting. \\
16 & Overall & Overall quality (H) & Overall, this is a high-quality sleep recommendation for this case. \\
\bottomrule
\end{tabular}
\end{table*}

\begin{table*}[htbp]

\centering
\scriptsize
\setlength{\tabcolsep}{4pt}
\renewcommand{\arraystretch}{1.12}
\caption{\textbf{Five-point score anchors for the Stage-2 rubric.}}
\label{tab:stage2-section-scales}
\label{tab:stage2-overall-scales}
\begin{tabular}{@{}>{\raggedright\arraybackslash}p{0.14\textwidth}
>{\raggedright\arraybackslash}p{0.80\textwidth}@{}}
\toprule
Scale & Guidance options \\
\midrule
A: Role & \textbf{1}: The section is mostly outside its role. \textbf{2}: It contains one important piece of out-of-role content. \textbf{3}: It contains several minor pieces of out-of-role content. \textbf{4}: It contains one minor piece of out-of-role content. \textbf{5}: Everything in it belongs to its role. \\
B: Consistency & \textbf{1}: Contradictions run through the section. \textbf{2}: It contains one important contradiction. \textbf{3}: It contains several minor inconsistencies. \textbf{4}: It contains one minor inconsistency. \textbf{5}: It is fully consistent. \\
C: Clarity & \textbf{1}: None of the important sleep data is presented clearly. \textbf{2}: Some is clear, but most important data is missing or unclear. \textbf{3}: About half of the important sleep data is clear. \textbf{4}: Most of the important sleep data is clear. \textbf{5}: All important sleep data is clear. \\
D: Actionability & \textbf{1}: Nothing can be acted on. \textbf{2}: An important part cannot be acted on. \textbf{3}: Several parts are too vague to act on. \textbf{4}: One part is slightly vague. \textbf{5}: Every part is specific and can be tracked. \\
\addlinespace[3pt]
E: Non-redundancy & \textbf{1}: The sections largely repeat each other. \textbf{2}: Two sections repeat each other in an important way. \textbf{3}: Two sections partly repeat each other. \textbf{4}: There is slight repetition between two sections. \textbf{5}: There is no repetition. \\
F: Safety & \textbf{1}: Clearly harmful content or explicit clinical or treatment advice is present. \textbf{2}: One important out-of-scope or potentially harmful item is present. \textbf{3}: Several minor out-of-scope items are present. \textbf{4}: One minor out-of-scope item is present, with no potential for harm. \textbf{5}: Nothing is out of scope or harmful. \\
G: Readability & \textbf{1}: Very difficult to read. \textbf{2}: Many errors affect readability. \textbf{3}: Some errors affect readability. \textbf{4}: A few errors are present, but the guidance remains easy to read. \textbf{5}: Easy to read, with no errors. \\
H: Overall quality & \textbf{1}: Very poor; it needs to be rewritten. \textbf{2}: Poor; it contains significant errors. \textbf{3}: Fair; it could be improved. \textbf{4}: Good; only slight improvements are possible. \textbf{5}: Excellent; no changes are needed. \\
\bottomrule
\end{tabular}
\end{table*}

\begin{figure*}[!t]
\centering
\includegraphics[width=0.8\textwidth]{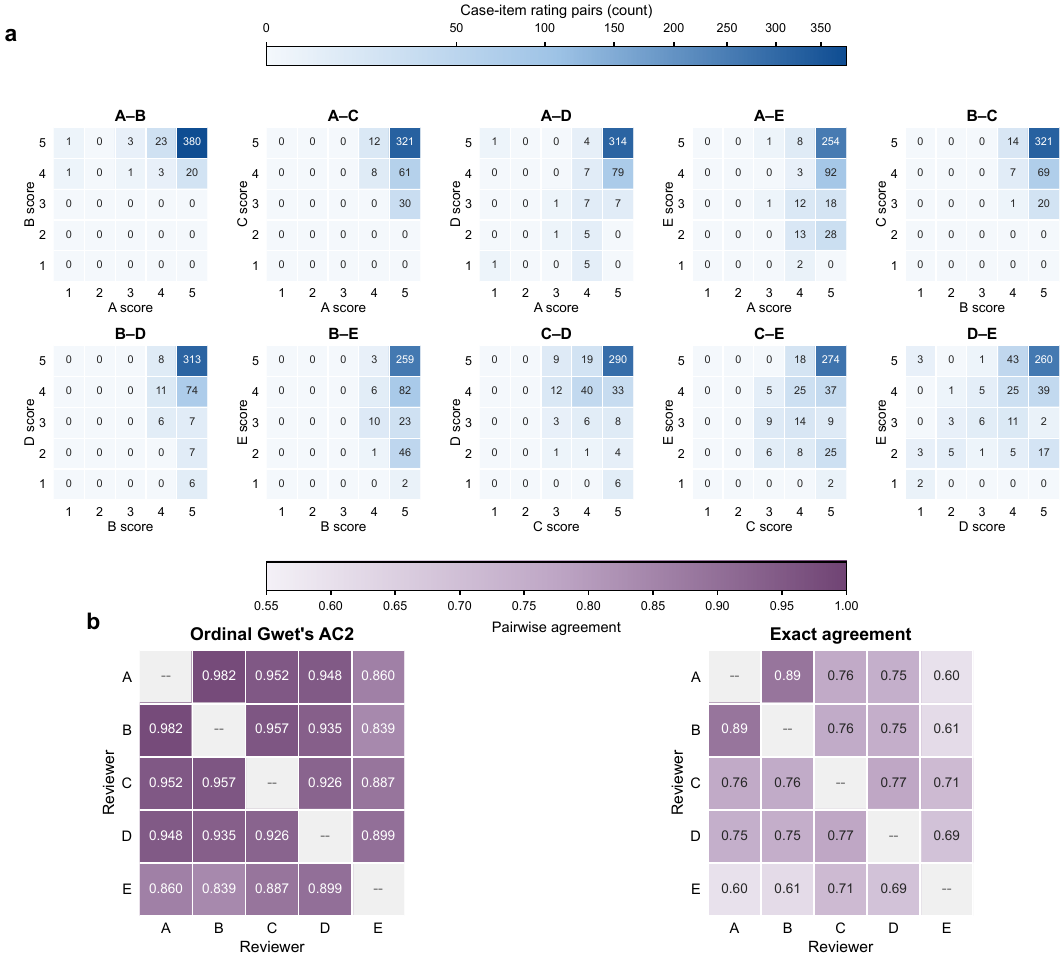}
\caption{\textbf{Pairwise agreement among five experienced MD-student
reviewers.} \textbf{a}, Pairwise $5\times5$ contingency matrices for all ten
reviewer pairs. Columns correspond to the first named reviewer and
rows to the second; cells report raw case--item counts. Each pair shared 27
cases, giving 432 case--item units. \textbf{b}, Full pairwise matrices of
ordinal Gwet's AC2 and exact agreement. Diagonal cells are omitted because
self-agreement is not an inter-rater comparison. Overall reliability estimates
are reported in Table~\ref{tab:stage2-agreement}.}
\label{fig:stage2-reviewer-agreement}
\end{figure*}

\paragraph{Administration and aggregation.}
The 90 cases were divided into ten nine-case groups and assigned to the ten
possible three-reviewer combinations drawn from five experienced MD students.
This balanced incomplete block design yielded three independent reviews per
case and 54 cases per reviewer. Reviewers judged only the displayed guidance,
without access to the participant record, evidence identifiers, evidence cards, reference guidance, or system identity. Item scores were summarized by
first taking the within-case median across three ratings and then averaging
across cases. Reliability was calculated across all 1,440 case--item units;
each case-level bootstrap sample retained all 16 items and their three ratings.
Figure~\ref{fig:stage2-reviewer-agreement} shows pairwise agreement among the
reviewers, and Table~\ref{tab:stage2-agreement} reports the overall reliability
estimates.

\clearpage
\onecolumn

\noindent
\begin{minipage}[t]{0.54\textwidth}
\vspace{0pt}
\centering
\captionof{table}{\textbf{Item-level Stage-2 human rubric scores.}
Means of within-case median scores across 90 cases for Qwen3.5-0.8B.}
\label{tab:stage2-rubric-full}

\scriptsize
\setlength{\tabcolsep}{3.5pt}
\renewcommand{\arraystretch}{0.90}

\begin{tabular}{@{}>{\raggedright\arraybackslash}p{0.76\linewidth}c@{}}
\toprule
Item & Mean score \\
\midrule
Summary role & 4.97 \\
Summary consistency & 5.00 \\
Summary clarity & 4.96 \\
Pattern Analysis role & 4.93 \\
Pattern Analysis consistency & 5.00 \\
Pattern Analysis clarity & 4.99 \\
Personalized Recommendation role & 4.69 \\
Personalized Recommendation consistency & 5.00 \\
Personalized Recommendation actionability & 4.39 \\
Follow-up role & 5.00 \\
Follow-up consistency & 5.00 \\
Follow-up actionability & 4.64 \\
Non-redundancy & 4.57 \\
Safety & 4.99 \\
Readability & 4.74 \\
Overall quality & 4.06 \\
\bottomrule
\end{tabular}
\end{minipage}
\hfill
\begin{minipage}[t]{0.42\textwidth}
\vspace{0pt}
\centering
\captionof{table}{\textbf{Inter-rater agreement for the Stage-2 human
evaluation.} Chance-corrected estimates include case-bootstrap 95\%
confidence intervals; observed agreement is reported descriptively.}
\label{tab:stage2-agreement}

\scriptsize
\renewcommand{\arraystretch}{1.10}

\begin{tabular*}{\linewidth}
{@{\extracolsep{\fill}}lc@{}}
\toprule
Measure & Result \\
\midrule

Ordinal Gwet's AC2 &
\begin{tabular}[c]{@{}c@{}}
0.922 \\[-2pt]
95\% CI: 0.912--0.932
\end{tabular}
\\[2pt]

Krippendorff's ordinal $\alpha$ &
\begin{tabular}[c]{@{}c@{}}
0.235 \\[-2pt]
95\% CI: 0.186--0.284
\end{tabular}
\\[2pt]

Exact agreement & 72.9\% \\

Difference $\leq 1$ point & 92.4\% \\

\bottomrule
\end{tabular*}
\end{minipage}

\subsection{Comparative Evaluation with the Wearable Record}
\label{app:human-compare}

Five raters (physicians and medical students) rated three systems on 30 test
cases: our $0.8$B model, GPT Health, and Guidance-only SFT ($4$B) given the full
text of all 22 evidence cards. For each case, the three outputs appeared in
random order under anonymized labels, and each output was rated by two raters.
Raters were given the participant's wearable record. The 16-item rubric of
Appendix~\ref{app:rubric-stage2} was extended with two items checked directly
against the record: \emph{data accuracy}, whether every statement about the
participant's data, such as values, dates, time periods, and changes over time,
matches the record; and \emph{use of the data}, whether the interpretations and
the reasons for the advice rest on what the record shows.
Table~\ref{tab:human-compare-full} reports per-case means over the 30 cases.
Among the individual items, Guidance-only SFT is rated higher on readability
($5.00$ vs.\ $4.85$), and safety is near the maximum for all three systems.

\noindent\begin{minipage}{\textwidth}
\centering
\captionof{table}{\textbf{Comparative human evaluation with the wearable
record available.} Values are per-case mean ratings (1--5) over 30 test cases.
The lower block reports four of the 16 original items individually.}
\label{tab:human-compare-full}
\scriptsize
\setlength{\tabcolsep}{6pt}
\renewcommand{\arraystretch}{1.0}
\begin{tabular}{@{}lccc@{}}
\toprule
& Ours & GPT & Guidance-only \\
Rubric item & (0.8B) & Health & SFT (4B) \\
\midrule
16 original items (mean) & 4.87 & 4.65 & 4.69 \\
Overall quality & 4.47 & 3.80 & 3.90 \\
Data accuracy & 4.98 & 4.08 & 4.67 \\
Use of the data & 4.88 & 4.43 & 4.28 \\
\addlinespace[2pt]
Safety & 5.00 & 4.90 & 4.98 \\
Readability & 4.85 & 4.28 & 5.00 \\
Non-redundancy & 4.48 & 4.35 & 3.95 \\
Recommendation actionability & 4.68 & 4.37 & 4.52 \\
\bottomrule
\end{tabular}
\end{minipage}

\paragraph{Inter-rater agreement.}
The two raters of each output gave the same score on $72.8\%$ of ratings and
scores within one point on $90.4\%$ (ordinal Gwet's AC2 of $0.94$;
Krippendorff's ordinal $\alpha$ of $0.23$).

\paragraph{Additional physicians.}
Two additional physicians each rated all three systems on 5 of the 30 cases.
Averaged over the two physicians, our model scored $4.64$ on the 16 original
items, compared with $4.41$ for GPT Health and $4.47$ for Guidance-only SFT;
$4.60$ on overall quality ($4.00$ and $3.80$); $4.80$ on data accuracy ($3.80$
and $4.70$); and $4.80$ on use of the data ($4.10$ and $4.20$).

\par\medskip

\clearpage

\nolinenumbers

\noindent
\begin{minipage}[t]{0.48\textwidth}
\vspace{0pt}

\section{Qualitative Examples}
\label{app:examples}

This section presents one complete Stage~1 evidence-grounded guidance
construction example and three representative Stage~2 input--output examples
generated by the Qwen3.5-0.8B system.

\subsection{Stage-1 Complete Evidence-Grounded
Guidance Construction Example}
\label{app:stage1-example}

\end{minipage}

\par\medskip

\begin{supplementbox}{Male, 61 years}
\small\raggedright
\textbf{Profile packet $P$ (display-compacted).}
Thirty-one valid nights were recorded from 29 July to 28 August 2023, with
100\% nightly coverage. Across the completed record, median sleep duration was
6~h 24~min. During the recent seven nights, median sleep duration was 6~h
8~min, compared with 6~h 24~min during the preceding seven nights. Recent
median time in bed was 6~h 55~min. Median time awake after falling asleep rose
from 48~min to 59~min, while median sleep efficiency decreased from 89.3\% to
87.1\%. \\
\medskip
\textbf{Observation $O$.}
Across the completed record, you usually slept about 6~h 24~min. Recently, you
typically slept about 6~h 8~min, compared with about 6~h 24~min during the
preceding period. You spent about 59~min awake after falling asleep recently,
compared with about 48~min previously. Sleep efficiency, the percentage of time
in bed spent asleep, was about 87.1\% recently versus 89.3\% previously. Recent
time in bed was about 6~h 55~min.\\ 
\medskip
\textbf{Selected evidence identifiers $E$.}
\nolinkurl{allow-more-time-for-sleep};
\nolinkurl{adults-18-to-64-usually-need-at-least-7-hours};
\nolinkurl{waso-quality-reference-for-adults-18-to-64};
\nolinkurl{persistent-sleep-difficulty-needs-clinical-context}.
\end{supplementbox}

\begin{supplementbox}{Evidence-card block $C$}
\footnotesize\raggedright
\textbf{Allow more time for sleep}
(\nolinkurl{allow-more-time-for-sleep}).
\textbf{Analysis.} When sleep is repeatedly short and the available time for
sleep is also limited, making somewhat more time available can increase sleep
duration, although the amount of change varies. \textbf{Recommendation.} When
practical, make a little more time for sleep on the days when it is repeatedly
short, for example by starting the sleep period a little earlier, without
forcing a fixed bedtime. \textbf{Follow-up.} After making more time available,
see whether sleep lasts longer across several nights. If it does, keep allowing
the additional time. \textbf{Select when.} Sleep duration is repeatedly short
recently and time in bed is also below 7~h in the same recent or weekday
pattern. Do not infer limited time available merely because sleep is below
7~h. \textbf{Action.} Make more time for sleep. \textbf{Do not claim.} Do not
promise a health, mood, or performance benefit; do not recommend an earlier
sleep start when time in bed is already at least 7~h; and do not turn this into
a fixed sleep schedule unless separate timing-consistency evidence is selected.

\medskip
\textbf{Sleep duration reference for adults aged 18 to 64}
(\nolinkurl{adults-18-to-64-usually-need-at-least-7-hours}).
\textbf{Analysis.} Adults generally need 7~h or more of sleep, although the
amount an individual needs can vary. \textbf{Select when.} The person is aged
18--64 and the repeated recent sleep duration is clearly below 7~h in a way
that changes the decision; interpret it alongside the earlier and whole-record
pattern. \textbf{Reference.} Usual lower sleep-duration reference for adults
aged 18--64: at least 420~min. \textbf{Do not claim.} Do not apply this card to
adults aged 65 or older; do not use it for one short night or when the
recommendation would be unchanged without the reference; and do not call 7~h
an exact personal requirement.

\medskip
\textbf{Time awake after sleep begins in adults}
(\nolinkurl{waso-quality-reference-for-adults-18-to-64}).
\textbf{Analysis.} Repeatedly spending longer awake after first falling asleep
can indicate interrupted sleep. For adults, 20~min or less is commonly used as
one general sign of good sleep quality, not as a diagnosis or an exact personal
target. \textbf{Select when.} Use this card for an adult younger than 65 when
time awake after falling asleep is repeatedly longer than 20~min across both
the whole record and recent nights, and this helps explain an ongoing
interruption pattern. \textbf{Reference.} Awake-time comparison for adults
younger than 65: at most 20~min. \textbf{Do not claim.} Do not apply this card
to adults aged 65 or older; do not diagnose insomnia or another sleep disorder;
do not use one night or a small isolated increase as the trigger; and describe
the measure in plain language as time awake after falling asleep.

\medskip
\textbf{Persistent sleep difficulty needs clinical context}
(\nolinkurl{persistent-sleep-difficulty-needs-clinical-context}).
\textbf{Analysis.} A repeated sleep pattern is more useful to discuss with a
clinician when the person also has ongoing trouble sleeping, feels unusually
sleepy during the day, or has trouble concentrating. The sleep record alone
cannot establish a disorder. \textbf{Recommendation.} If you are also having
ongoing sleep difficulty, unusual daytime sleepiness, or trouble concentrating,
discuss the repeated pattern with a clinician. \textbf{Follow-up.} Before the
discussion, note which of those problems you notice and how often they occur.
\textbf{Select when.} A prominent repeated duration or continuity problem
remains unresolved, no selected direct self-directed step addresses it, and
clinical care remains explicitly conditional on unrecorded ongoing trouble
sleeping, unusual daytime sleepiness, or trouble concentrating.
\textbf{Action.} Discuss persistent sleep difficulty. \textbf{Only when.} If
you are also having ongoing sleep difficulty, unusual daytime sleepiness, or
trouble concentrating. \textbf{Do not claim.} Do not diagnose insomnia or
infer daytime impairment; do not use professional care as a generic closing
sentence; and do not recommend medication or a treatment protocol.
\end{supplementbox}

\begin{supplementbox}{Four-section guidance}
\small\raggedright
\textbf{Summary.} You usually slept about 6~h 24~min across the 31-night
record, below 7~h. During the recent seven nights, you typically slept about
6~h 8~min, compared with about 6~h 24~min during the preceding seven nights.
Recent time in bed was about 6~h 55~min. You spent more time awake after
falling asleep recently---about 59~min, compared with about 48~min during the
preceding period.

\medskip
\textbf{Pattern Analysis.} Recent sleep was shorter. Time awake after falling
asleep also remained elevated, suggesting that sleep was more interrupted. For
adults, 20~min or less is commonly used as one general sign of good sleep
quality, not as a diagnosis or an exact personal target.

\medskip
\textbf{Personalized Recommendation.} \\
\textbf{Make more time for sleep.}
Recent sleep lasted about 6~h 8~min, and time in bed was about 6~h 55~min, so
both were short. When practical, allow a little more time for sleep on days
when it is repeatedly short, such as by starting the sleep period a little
earlier, without forcing a fixed bedtime.

\textbf{Discuss persistent sleep difficulty.} You spent more time awake after
falling asleep recently, and sleep efficiency was lower than during the
preceding period. If you also have ongoing sleep difficulty, unusual daytime
sleepiness, or trouble concentrating, discuss this repeated pattern with a
clinician.

\medskip
\textbf{Follow-up.} After making more time available, assess over the next
several nights whether sleep lasts longer. If it does, continue allowing the
additional time. If you notice ongoing sleep difficulty or unusual daytime
sleepiness, note how often it occurs and discuss the repeated pattern with a
clinician.
\end{supplementbox}

\par\medskip
\noindent
\begin{minipage}[t]{0.48\textwidth}

\subsection{Stage-2 Representative Input--Output Examples}
\label{app:stage2-examples}

We present three representative examples generated by Qwen3.5-0.8B, spanning
distinct sleep patterns. Each example shows the input profile packet and the
corresponding four-section guidance.

\subsubsection{Stable Recent Sleep Pattern}

\end{minipage}

\par\medskip

\begin{supplementbox}{Female, 41 years}
\small\raggedright
\textbf{Profile packet $P$ (display-compacted).}
Nineteen valid nights were recorded from 17 March to 4 April 2023, with 100\%
nightly coverage. Across the completed record, median sleep duration was 8~h
8~min. During the recent seven nights, median sleep duration was 7~h 57~min,
compared with 8~h 27~min during the preceding seven nights. Recent median sleep
efficiency was 90.8\%, compared with 89.5\% previously, and median time awake
after falling asleep decreased from 1~h 5~min to 41~min.
\end{supplementbox}

\begin{supplementbox}{Four-section guidance}
\small\raggedright
\textbf{Summary.} Across the completed record, you usually slept about 8~h
8~min. During the recent seven nights, you usually slept about 7~h 57~min,
30~min less than during the preceding seven nights. Sleep efficiency was 90.8\%
recently, compared with 89.5\% previously. You spent 41~min awake after falling
asleep recently, down from 1~h 5~min.

\medskip
\textbf{Pattern Analysis.} Your recent sleep duration remained adequate despite
being shorter than during the preceding seven nights. Sleep efficiency improved,
and you spent less time awake after falling asleep. Together, these changes
support maintaining the recent pattern.

\medskip
\textbf{Personalized Recommendation.} \\
\textbf{Protect the current sleep
duration.} Your recent sleep duration remained adequate, and sleep efficiency
improved. Continue leaving a similar amount of time for sleep.

\medskip
\textbf{Follow-up.} Over the next several nights, assess whether your sleep
duration remains near its recent level. If it does, continue leaving a similar
amount of time for sleep.
\end{supplementbox}

\par\medskip
\noindent
\begin{minipage}[t]{0.48\textwidth}

\subsubsection{Weekday--Weekend Timing Shift}

\end{minipage}
\par\medskip

\begin{supplementbox}{Female, 52 years}
\small\raggedright
\textbf{Profile packet $P$ (display-compacted).}
Twenty-one valid nights were recorded from 7 to 27 January 2020, with 100\%
nightly coverage. Sleep duration was similar on weekdays and weekends. Sleep
usually began at about 10:46~p.m. on weekdays and 11:22~p.m. on weekends. Final
waking occurred at about 6:46~a.m. on weekdays and 8:03~a.m. on weekends. The
midpoint of sleep shifted from about 2:47~a.m. on weekdays to 3:50~a.m. on
weekends.
\end{supplementbox}

\begin{supplementbox}{Four-section guidance}
\small\raggedright
\textbf{Summary.} Across the record, your sleep duration was generally steady.
On weekends, you usually fell asleep later and woke later than on weekdays.
The midpoint of sleep also occurred later on weekends, showing that the entire
sleep period shifted later, while sleep duration remained similar.

\medskip
\textbf{Pattern Analysis.} Your sleep period shifted later on weekends, but
sleep duration stayed similar. Regular sleep timing is one dimension of sleep
health, but no single schedule or amount of variation is right for everyone.
Clock times alone cannot establish internal circadian timing or a circadian
disorder.

\medskip
\textbf{Personalized Recommendation.} \\
\textbf{Make sleep timing more
consistent.} Your sleep began and ended later on weekends, while sleep duration
remained similar. When practical, bring weekend sleep and wake times closer to
weekday timing while still leaving enough time to sleep.

\medskip
\textbf{Follow-up.} Over the next several weekends, assess whether weekend sleep
and wake times move closer to weekday timing without shortening sleep. If they
do, maintain the more consistent timing.
\end{supplementbox}

\par\medskip
\noindent
\begin{minipage}[t]{0.48\textwidth}

\subsubsection{Persistent Sleep Interruption with a Conditional Behavioral Step}

\end{minipage}
\par\medskip

\begin{supplementbox}{Male, 70 years}
\small\raggedright
\textbf{Profile packet $P$ (display-compacted).}
Fifteen valid nights were recorded from 3 to 17 October 2023, with 100\%
nightly coverage. Across the completed record, median sleep duration was 6~h
56~min and median time awake after falling asleep was 55~min. During the recent
seven nights, median sleep duration was 6~h 54~min, compared with 7~h 8~min
previously. Recent median time in bed was 7~h 49~min, and median time awake after
falling asleep increased from 48~min to 1~h 1~min.
\end{supplementbox}

\begin{supplementbox}{Four-section guidance}
\small\raggedright
\textbf{Summary.} Across the completed record, you usually slept about 6~h
56~min and spent about 55~min awake after falling asleep. Recently, you slept
about 6~h 54~min, compared with about 7~h 8~min during the preceding period.
You recently spent about 1~h 1~min awake after falling asleep, up from about
48~min, while time in bed was about 7~h 49~min.

\medskip
\textbf{Pattern Analysis.} You recently spent about 1~h 1~min awake after
falling asleep, suggesting that sleep was more interrupted. For older adults,
30~min or less is commonly used as one general sign of good sleep quality, not
as a diagnosis or exact personal target. Because recent time in bed was about
7~h 49~min, limited time in bed was not the main constraint.

\medskip
\textbf{Personalized Recommendation.} \\
\textbf{Move caffeine earlier if
relevant.} Recent sleep lasted about 6~h 54~min, and you spent about 1~h 1~min
awake after falling asleep. If you use caffeine near bedtime, try having it
earlier or avoiding it near bedtime.

\medskip
\textbf{Follow-up.} After moving caffeine earlier, assess across several nights
whether you sleep longer and spend less time awake after falling asleep. If
both improve, keep the earlier timing; if neither changes, caffeine timing may
not be an important contributor to this pattern.
\end{supplementbox}

\section{Composer Prompt}
\label{app:composer-prompt}

The Composer prompt consists of a shared writing contract and four
section-specific instruction blocks. The print-oriented version below preserves
the substantive generation, evidence-use, and reader-facing requirements while
omitting implementation-specific schemas, grounding metadata, and retry
instructions.

\centering
\captionof{table}{\textbf{Prompt for the Composer Agent.}}
\label{tab:composer-prompt}

\begin{supplementbox}{Prompt for the Composer Agent}
\footnotesize
\raggedright
\sloppy
\setlength{\emergencystretch}{2em}
\textbf{Role.} Write one connected sleep recommendation for the person represented by
the input. Use the profile packet $P$ for personal facts, verify the supplied retrieval
focus against those facts, and use only the selected evidence cards $C$ for general
interpretation and guidance.

\smallskip
\textbf{Output.} Return exactly four sections in this order: \emph{Summary},
\emph{Pattern Analysis}, \emph{Personalized Recommendation}, and \emph{Follow-up}.
Each section must advance the same reasoning chain without repeating information that is
already clear.

\smallskip
\textbf{Summary.} Give a selective longitudinal account of the record. Begin with the
usual sleep pattern, then describe the recent change or comparison that matters most.
Distinguish sleep duration from time in bed when their difference changes the
interpretation. Mention the latest night or a dated fluctuation only when it clarifies
the trajectory. State observations only, without causes, diagnoses, evidence
interpretation, or advice.

\smallskip
\textbf{Pattern Analysis.} Interpret the main repeated pattern and explain why it
matters. Include the strongest personal fact that limits an overly simple explanation.
Apply the selected evidence to the case rather than describing the source itself. Use a
population reference only when it changes the interpretation, and present it as general
context rather than a diagnosis or exact personal target. Do not give instructions in
this section.

\smallskip
\textbf{Personalized Recommendation.} Give the complete set of supported next steps as
one to four short, action-titled items. Each item should first state the personal finding
that makes the action relevant and then state a concrete, feasible action. Prefer actions
supported by observed data. If a behavior or symptom was not recorded, make the action
explicitly conditional. Do not invent a dose, schedule, numeric target, diagnosis,
preference, causal explanation, or promised benefit.

\smallskip
\textbf{Follow-up.} Explain how to evaluate the principal recommendation across a
comparable block of nights. Name the sleep outcome to watch and, when useful, its current
multi-night value. State what a repeated change would mean. Do not infer success from one
night, introduce a new intervention, or add professional care unless the selected
evidence supports a conditional clinical boundary.

\smallskip
\textbf{Writing and grounding.} Write directly and supportively in plain English. Use
readable dates, clock times, durations, and percentages. Preserve all case values,
conditions, limitations, and evidence boundaries. Do not expose metric identifiers,
evidence identifiers, retrieval steps, audit language, statistical jargon, or the data
collection technology in reader-facing text. Use only evidence present in $C$ and retain
only evidence identifiers that resolve in the offline Evidence Bank.
\end{supplementbox}

\end{document}